\documentclass[authoryear,review,12pt]{elsarticle}
\usepackage{algorithm}

\usepackage[breaklinks]{hyperref}	
\usepackage{graphicx}
\usepackage{times}
\usepackage{amsmath,amssymb}
\usepackage{makeidx}  
\usepackage{amsmath,amsfonts,amssymb,dsfont}
\usepackage{pdfpages,graphics,epsfig}
\usepackage{array}
\usepackage{balance}
\usepackage{tabularx}
\usepackage{longtable,multirow}
\usepackage{caption}
\usepackage{subcaption}
\usepackage{algorithmic}

\usepackage{amssymb}
\usepackage{latexsym}

\usepackage{xcolor}
\definecolor{newcolor}{rgb}{.8,.349,.1}
\usepackage{todonotes}

\usepackage{ifpdf}
\ifpdf
 \usepackage{graphicx}
 \graphicspath{{./Images/}}
 \DeclareGraphicsExtensions{.pdf,.png}
\else
 \usepackage[dvips]{graphicx}
 \graphicspath{{./Images/}}
 \DeclareGraphicsExtensions{.eps,.png}
\fi

\newcolumntype{H}{>{\setbox0=\hbox\bgroup}c<{\egroup}@{}}
\renewcommand{\v}[1]{\mathbf{#1}}

\begin{document}
\begin{frontmatter}
\title{Probabilistic classifiers with low rank indefinite kernels}
\author{Frank-Michael Schleif\corref{cor1}$^1$}
\cortext[cor1]{Corresponding author}
\ead{schleify@cs.bham.ac.uk}
\author{Andrej Gisbrecht$^2$}
\ead{andrej.gisbrecht@aalto.fi}
\author{Peter Tino$^1$}
\ead{pxt@cs.bham.ac.uk}
\address{$^1$School of Computer Science, University of Birmingham, Birmingham B15 2TT, UK}
\address{$^2$Helsinki Institute for Information Technology,\\Department of Computer Science, Aalto University, Finland}
\newtheorem{definition}{Definition}{\bfseries}{\rmfamily}
\begin{abstract}
	Indefinite similarity measures can be  frequently found in bio-informatics by means of alignment scores,
	but are also common in other fields like shape measures in image retrieval.
	Lacking an underlying vector space, the data are given as pairwise similarities only. The few algorithms
	available for such data do not scale to larger datasets. Focusing on probabilistic batch classifiers, the Indefinite Kernel Fisher
	Discriminant (iKFD) and the Probabilistic Classification Vector Machine (PCVM) are both
         effective algorithms for this type of  data but, with cubic complexity. 
	Here we propose an extension of iKFD and PCVM such that linear runtime and memory complexity is achieved for low
	rank indefinite kernels. Employing the Nystr\"om approximation for indefinite kernels, we also propose a new
	almost parameter free approach to identify the landmarks, restricted to a supervised learning problem.
	Evaluations at several larger similarity data from various domains show that the proposed 
	methods provides similar generalization capabilities while being  easier to parametrize  and substantially faster for large scale data.
\end{abstract}


\end{frontmatter}

\section{Introduction}
Domain specific proximity measures, like alignment scores in bioinformatics \cite{citeulike:668527}, the modified Hausdorff-distance for
structural pattern recognition \cite{576361}, shape retrieval measures like the inner distance \cite{DBLP:journals/pami/LingJ07}
and many other ones generate non-metric or indefinite similarities or dissimilarities. Classical learning algorithms like kernel
machines assume Euclidean metric properties in the underlying data space and may not be applicable for this type of data.

Only few machine learning methods have been proposed for non-metric proximity data, like the indefinite kernel Fisher discriminant (iKFD) \cite{Haasdonk2008,Pekalska20091017}, the
probabilistic classification vector machine (PCVM) \cite{DBLP:journals/tnn/ChenTY09} or the indefinite Support Vector Machine (iSVM) in different formulations \cite{Haasdonk2005482,DBLP:conf/acml/AlabdulmohsinGZ14,7254195}.
For the PCVM the provided kernel evaluations are considered only as basis functions and no mercer conditions are implied.
In contrast to the iKFD the PCVM is a \emph{sparse} probabilistic kernel classifier pruning unused basis functions during training, applicable to arbitrary positive 
definite \emph{and} indefinite kernel matrices. A recent review about learning with indefinite proximities can be found in \cite{Schleif2015f}.

While being very efficient these methods do not scale to larger datasets with in general cubic complexity. In \cite{Schleif2015d,Schleif2015g} 
the authors proposed a few Nystr\"om based (see e.g. \cite{DBLP:conf/nips/WilliamsS00}) approximation techniques to improve the scalability 
of the PCVM for low rank matrices. The suggested techniques use the Nystr\"om approximation in a non-trivial
way to provide \emph{exact} eigenvalue estimations also for \emph{indefinite} kernel matrices.  This approach is very 
generic and can be applied in different algorithms. In this contribution we further extend our previous work and
not only derive a low rank approximation of the indefinite kernel Fisher discriminant, but also address the landmark selection
from a novel view point.  The obtained Ny-iKFD approach is linear in runtime and memory consumption for 
low rank matrices. The formulation is exact if the rank of the matrix equals the number of independent landmarks points.
The selection of the landmarks of the Nystr\"om approximation is a critical point addressed in  previous work 
(see e.g. \cite{DBLP:journals/tnn/ZhangK10a,DBLP:conf/icml/SiHD14,DeBrabanter20101484}). 
In general these strategies use the full psd kernel matrix or expect that the kernel is of some standard class like an 
RBF kernel. In each case the approaches presented so far are costly in runtime and memory consumption as can
be seen in the subsequent experiments.

Additionally, former approaches for landmark selection aim on generic matrix
reconstructions of positive semi definite (psd) kernels. We propose a restricted reconstruction of the psd or non-psd 
kernel matrix with respect to a \emph{supervised} learning scenario only. We do not any longer expect to obtain an
accurate kernel reconstruction from the approximated matrix (e.g. by using the Frobenius norm) but are pleased
if the approximated matrix preserves the class boundaries in the data space.

In \cite{Schleif2015g} the authors derived methods to approximate large proximity matrices by means of
the Nystr\"om approximation and conversion rules between similarities and dissimilarities. These techniques have been
applied in \cite{Schleif2015d} and \cite{Schleif2015j} in a proof of concept setting, to obtain approximate models 
for the Probabilistic Classification Vector Machine and the Indefinite Fisher Kernel Discriminant analysis 
using a random landmark selection scheme.  This work is substantially extended and detailed in this article with a specific focus 
on \emph{indefinite} kernels, only. A novel landmark selection scheme is proposed. Based on this new landmark selection 
scheme we provide detailed new experimental  results and compare to alternative landmark selection approaches.

Structure of the paper: First we give some basic notations necessary in the subsequent derivations. Then we review iKFD and PCVM 
as well as some approximation concepts proposed by the authors in \cite{Schleif2015d} which are based on the well known
Nystr\"om approximation. Subsequently, we consider the landmark selection problem in more detail and show empirically results motivating a
supervised selection strategy. Finally we detail the reformulation of iKFD and PCVM based on the introduced concepts and show the  
efficiency in comparison to Ny-PCVM and Ny-iKFD for various indefinite proximity benchmark data sets.

\section{Methods}
\subsection{Notation and basic concepts}
Let $\mathcal{X}$ be a collection of $N$ objects $x_i$, $i=1,2,...,N$,  in some input space. 
If the similarity function or inner product used to compare two objects $x_i$, $x_j$ is metric,
proper mercer kernels can be obtained as discussed subsequently. A classical similarity function 
in this context, is the  Euclidean inner product with the respective Euclidean distance which is a 
frequent core component of various metric kernel functions, like the famous radial basis function (rbf) kernel.

Now, let  $\phi: \mathcal{X} \mapsto \mathcal{H}$ be a mapping of patterns from $\mathcal{X}$
to a high-dimensional or infinite dimensional Hilbert space $\mathcal{H}$ equipped with
the inner product $\langle \cdot, \cdot \rangle_\mathcal{H}$. The transformation $\phi$ 
is in general a non-linear mapping to a high-dimensional space $\mathcal{H}$ and may
in general not be given in an explicit form. Instead a kernel function $k: \mathcal{X} \times \mathcal{X}
\mapsto \mathbb{R}$ is given which encodes the inner product in $\mathcal{H}$.
The kernel $k$ is a positive (semi) definite function such that $k(x,x^\prime)  =\phi(x)^\top \phi(x^\prime)$
 for any $x, x^\prime \in \mathcal{X}$. The matrix $K:=\Phi^\top \Phi$ is an $N \times N$ kernel matrix
derived from the training data, where  $\Phi:[\phi(x_1),\hdots, \phi(x_N)]$ is a matrix of images (column vectors) of
the training data in $\mathcal{H}$.
The motivation for such an embedding comes with the hope that the non-linear transformation of  input data into higher dimensional $\mathcal{H}$ allows for using linear techniques in $\mathcal{H}$.
Kernelized methods process the 
embedded data points in a feature space utilizing only the inner products $\langle \cdot,\cdot \rangle_{\mathcal{H}}$ 
(kernel trick) \citep{Cristianini2004a}, without the need to explicitly calculate $\phi$. 
The specific kernel function can be
very generic. Most prominent are the linear kernel with $k(\mathbf{x},\mathbf{x^\prime})=\langle \phi(\mathbf{x}) , \phi(\mathbf{x^\prime}) \rangle$
where $\langle \phi(\mathbf{x}) , \phi(\mathbf{x^\prime}) \rangle$ is the Euclidean inner product or the 
rbf kernel $k(\mathbf{x},\mathbf{x^\prime})=\exp{\left (-\frac{||\mathbf{x}-\mathbf{x^\prime}||^2}{2\sigma^2}\right)}$,
with $\sigma$ as a free parameter. Thereby it is assumed that the kernel function $k(\mathbf{x},\mathbf{x^\prime})$ is positive semi definite (psd). 
This assumption is however not always fulfilled, and the underlying similarity measure may not be metric and hence not lead
to a mercer kernel. Examples can be easily found in domain specific similarity measures as mentioned before and detailed later on. These measures imply \emph{indefinite}
kernels. In what follows we will review some basic concepts and approaches related to such non-metric situations.

\subsection{Krein and Pseudo-Euclidean spaces}
A Krein space is an \emph{indefinite} inner product space endowed with a Hilbertian topology.
\begin{definition}[Inner products and inner product space]
Let $\mathcal{K}$ be a real vector space.  An inner product space with an indefinite inner product
$\langle \cdot, \cdot \rangle_\mathcal{K}$ on $\mathcal{K}$ is a bi-linear form where all $f,g,h \in \mathcal{K}$ and $\alpha \in \mathbb{R}$ obey the
following conditions. 
\begin{itemize}
	\item Symmetry: $\langle f,g \rangle_\mathcal{K} = \langle g,f \rangle_\mathcal{K}$
	\item linearity:  $\langle \alpha f + g,h \rangle_\mathcal{K} = \alpha \langle f,h \rangle_\mathcal{K} + \langle g,h \rangle_\mathcal{K}$;
	\item $ \langle f, g \rangle_\mathcal{K} = 0 \; \forall g \in K$ implies $ f = 0$
\end{itemize}
\end{definition}
An inner product is positive definite
if $\forall f \in \mathcal{K}$,  $ \langle f,  f \rangle_\mathcal{K} \ge 0$, negative definite if $\forall f \in \mathcal{K}$, $\langle f, f \rangle_\mathcal{K} \le 0$,
otherwise it is indefinite. A vector space $\mathcal{K}$ with inner product $\langle \cdot, \cdot \rangle_\mathcal{K}$ is called an inner product space.
\begin{definition}[Krein space and pseudo Euclidean space]\label{def:kreinspace}
An inner product space $(\mathcal{K}, \langle \cdot, \cdot \rangle_\mathcal{K})$ is a Krein space if we have two Hilbert spaces $\mathcal{H}_+$
and $\mathcal{H}_-$ spanning $\mathcal{K}$ such that $\forall f \in \mathcal{K}$ we have $f = f_+ + f_-$ with $f_+ \in \mathcal{H}_+$
and $f_- \in \mathcal{H}_-$ and 
$\forall f,g \in \mathcal{K}$, $ \langle f,g \rangle_\mathcal{K} = \langle f_+,g_+ \rangle_{\mathcal{H}_+} - \langle f_-,g_- \rangle_{\mathcal{H}_-}$.
A finite-dimensional Krein-space is a so called pseudo Euclidean space (pE).
\end{definition}

Indefinite kernels are typically observed by means of domain specific
non-metric similarity functions (such as alignment functions used in biology \citep{citeulike:668527}), by
specific kernel functions - e.g.  the Manhattan kernel $k(\mathbf{x},\mathbf{x^\prime})= - ||\mathbf{x}-\mathbf{x^\prime}||_1$,
tangent distance kernel \citep{DBLP:conf/icpr/HaasdonkK02} or divergence measures plugged into standard kernel
functions \citep{Cichocki20101532}. Another source of non-psd kernels are noise artifacts on standard kernel functions \citep{Haasdonk2005482}.

For such spaces vectors can have negative squared "norm",  negative squared "distances" and the concept of orthogonality is different
from the usual Euclidean case. In the subsequent experiments our input data are in general given by a symmetric indefinite kernel matrix ${K}$.

Given a symmetric \emph{dissimilarity} matrix with zero diagonal \footnote{A similarity matrix can be easily converted into squared dissimilarities
using $d^2(x,y) = k(x,x)+k(y,y)-2\cdot k(x,y)$.}, an embedding of the data in a pseudo-Euclidean vector space
determined by the eigenvector decomposition of the associated similarity matrix $\mathbf{S}$ is always possible \citep{Goldfarb1984575}
\footnote{
The associated similarity matrix can be obtained by double centering \citep{Pekalska2005a} of the (squared) dissimilarity matrix.
$\mathbf{S} = -\mathbf{J} \mathbf{D} \mathbf{J}/2$  with $\mathbf{J} = (\mathbf{I}-\mathbf{1}\mathbf{1}^\top/N)$
and identity matrix $\mathbf{I}$  and vector of ones $\mathbf{1}$.}

Given the eigendecomposition of  $\mathbf{S} $,  $\mathbf{S} = \mathbf{U} \mathbf{\Lambda} \mathbf{U}^\top$,
we can compute the corresponding vectorial representation $\mathbf{V}$
in the pseudo-Euclidean space by
\begin{equation}
\mathbf{V} = \mathbf{U}_{p+q+z} \left|\mathbf{\Lambda}_{p+q+z}\right|^{1/2}
\label{eq:embedding}
\end{equation}
where $\mathbf{\Lambda}_{p+q+z}$ consists of $p$ positive, $q$ negative non-zero eigenvalues and $z$ zero eigenvalues.
 $\mathbf{U}_{p+q+z}$ consists of the corresponding eigenvectors. The 
triplet $(p,q,z)$ is also referred to  as the signature of the
Pseudo-Euclidean space.
This operation is however very costly and should be avoided for larger data sets. 
A detailed presentation of similarity and dissimilarity measures, 
and mathematical aspects of metric and non-metric spaces is provided in \citep{Pekalska2005a}.

\subsection{Indefinite Fisher and kernel quadratic discriminant}
In  \cite{Haasdonk2008,Pekalska20091017} the indefinite
kernel Fisher discriminant analysis (iKFD) and indefinite kernel quadratic discriminant analysis (iKQD)
was proposed focusing on binary classification problems, recently extended by a weighting scheme in \cite{Yang20131}\footnote{For multiclass problems a classical 1 vs rest wrapper is used within this paper}.

The initial idea is to embed the training data into a Krein space (see Def. \ref{def:kreinspace}) and to apply a modified  kernel Fisher discriminant analysis 
or kernel quadratic discriminant analysis for indefinite kernels. 
Consider binary classification and a data set of input-target training pairs
$D=\{\mathbf{x}_{i},y_{i}\}_{i=1}^{N}$, where $y_{i}\in \{-1,+1\}$. 
Given the indefinite kernel matrix $K$ and the embedded data in a pseudo-Euclidean space (pE), 
the linear Fisher Discriminant function $f(x)=\langle w,\Phi(\mathbf{x})\rangle_{pE} + b$ is based on a weight vector $\v{w}$
such that the between-class scatter is maximized while the within-class scatter is minimized along $w$. 
$\Phi (\mathbf{x})$ is a vector of basis function evaluations for data item $\mathbf{x}$ and $b$ is a bias term.
This direction is obtained by 
maximizing the Fisher criterion in the pseudo Euclidean space
\[
	J(\v{w}) = \frac{{\langle \v{w} ,\Sigma^b_{pE} \v{w} \rangle}_{pE}}{{\langle \v{w} ,\Sigma^w_{pE} \v{w}\rangle}_{pE}}
\]
where $\Sigma^b_{pE} = \Sigma_b J $ is the scatter matrix in the pseudo Euclidean space, with $J = diag(\mathbf{1}_p, -\mathbf{1}_q)$ where $\mathbf{1}_p \in \mathbb{R}^p$ denotes the $p$-dimensional vector of all ones.
The number of positive eigenvalues is denoted by $p$ and for the negative eigenvalues by $q$.
The \emph{within-scatter-matrix} in the pseudo-Euclidean space is given as $\Sigma^w_{pE} = \Sigma_w J$. 
The \emph{Euclidean} between- and within-scatter-matrices can be expressed as:
\begin{eqnarray}
	\Sigma_{b} &=& (\mu_+ - \mu_-)(\mu_+ - \mu_-)^\top\label{eq:between_scatter}\\
	\Sigma_{w} &=& \sum_{i \in I_+} (\phi(\mathbf{x}_{i}) - \mu_+)(\phi(\mathbf{x}_{i}) - \mu_+)^\top + \sum_{i \in I_-} (\phi(\mathbf{x}_{i}) - \mu_-)(\phi(\mathbf{x}_{i}) - \mu_-)^\top  \label{eq:within_scatter}
\end{eqnarray}
Where the set of indices of each class are $I_+ := \{i : y_i = +1\}$ and $I_- := \{i : y_i = -1\}$. 
In \cite{Haasdonk2008} it is shown that the Fisher Discriminant in
the pE space $\in \mathbb{R}^{(p,q)}$ is identical to the Fisher Discriminant in the associated
Euclidean space $\mathbb{R}^{p+q}$. To avoid the explicit embedding into the pE space a kernelization 
is considered such that the weight vector $w \in \mathbb{R}^{(p,q)}$ 
is expressed as a linear combination of the training data $\phi(\mathbf{x}_{i}$, hence $w = \sum_{i=1}^N \alpha_i \phi(\mathbf{x}_{i}$.
Transferred to the Fisher criterion this allows to use the kernel trick.
A similar strategy can be used for KQD as well as  the indefinite kernel PCA \cite{Pekalska20091017}.  

\subsection{Probabilistic Classification Vector Learning}
PCVM uses a kernel regression model $ \sum_{i=1}^{N} w_{i} \phi _{i}(\mathbf{x})+b $
with a link function, with $w_{i}$ being again the weights of the basis functions $\phi _{i}(\mathbf{x})$
and $b$ as a bias term. 
The Expectation Maximization (EM)
implementation of PCVM \cite{Chen2014356} uses the probit link function, i.e.
$
\Psi (x)=\int_{-\infty }^{x}\mathcal{N}(t|0,1)dt\text{,}
$
where $\Psi (x)$ is the cumulative distribution of the normal distribution $\mathcal{N}(0,1)$. 
We get: %
$
l(\mathbf{x};\mathbf{w},b)=\Psi \left( \sum_{i=1}^{N} w_{i} \phi _{i}(\mathbf{x})+b \right) = \Psi \left ( \Phi (\mathbf{x}) \mathbf{w} +b \right ) 
\label{model}
$

In the PCVM formulation \cite{DBLP:journals/tnn/ChenTY09}, a truncated Gaussian prior $N_t$  with support on $[0,\infty)$ and mode at $0$
is introduced for each weight $w_{i}$ and a zero-mean Gaussian prior is adopted
for the bias $b$. The priors are assumed to be mutually independent.
$
       p(\mathbf{w|\alpha }) 			=\prod\limits_{i=1}^{N}p(w_{i}\mathbf{|}\alpha_{i})=\prod\limits_{i=1}^{N}N_{t}(w_{i}|0,\alpha _{i}^{-1})\text{,} \quad
       p(b\mathbf{|}\beta)             		= \mathcal{N}(b|0,\beta^{-1})\text{,}
$
$\,\delta (x)= \mathbf{1}, \; {x> 0}$.
\begin{eqnarray}
p(w_{i}|\alpha _{i}) &=&\left\{
\begin{array}{ll}
2\mathcal{N}(w_{i}\mathbf{|}0,\alpha _{i}^{-1}) & \mbox{if $y_i w_{i}> 0$} \\
0 & \mbox{otherwise}%
\end{array}%
\right.  \notag 
=
2\mathcal{N}(w_{i}\mathbf{|}0,\alpha _{i}^{-1})\cdot \delta (y_i w_{i}).
\label{trunct_prior}
\end{eqnarray}
We follow the standard probabilistic formulation and assume that $z(\mathbf{x})= \Phi (\mathbf{x}) \mathbf{w} +b $
is corrupted by an additive random noise $\epsilon$ , where $\epsilon \sim \mathcal{N}(0, 1)$. According to the probit link model,
we have:
\begin{eqnarray}\label{eq:decision_function}
	h (\mathbf{x})=\Phi (\mathbf{x}) \mathbf{w} +b +\epsilon \ge 0, y=1,\quad
         h (\mathbf{x}) = \Phi (\mathbf{x}) \mathbf{w} +b +\epsilon <0, y=-1 
\end{eqnarray}
and obtain:
$
 p(y= 1|\mathbf{x}, \mathbf{w},b) = p( \Phi (\mathbf{x}) \mathbf{w} +b +\epsilon \ge 0) = \Psi( \Phi (\mathbf{x}) \mathbf{w} +b).
$
$h (\mathbf{x})$ is a latent variable because $\epsilon$ is an unobservable variable. We collect evaluations of $h (\mathbf{x})$
at training points in a vector $\mathbf{H}(\mathbf{x}) = (h (\mathbf{x_1}),\hdots,h (\mathbf{x_N}))^\top$.
In the expectation step the expected value $\mathbf{\bar{H}} $ of $\mathbf{H}$ with 
respect to the posterior distribution over the latent variables is calculated (given old values $\mathbf{w}^\text{old},b^\text{old}$).
In the maximization step the parameters are updated through 

\begin{eqnarray}
	\mathbf{w}^\text{new} &=& M {(M \Phi^\top (\mathbf{x}) \Phi (\mathbf{x}) M + I_N)}^{-1}
  M (\Phi^\top (\mathbf{x}) \mathbf{\bar{H}} - b \Phi^\top (\mathbf{x})\mathbf{I})\label{eq:weights_update}\\
	\mathbf{b}^\text{new}  &=& t (1+t N t)^{-1} t (\mathbf{I}^\top   \mathbf{\bar{H}}- \mathbf{I}^\top \Phi (\mathbf{x}) \mathbf{w}) \label{eq:bias_update}
\end{eqnarray}
where $I_N$ is a N-dimensional identity matrix and  $\mathbf{I}$ a all-ones vector, the diagonal elements in the diagonal matrix $M$ are:
\begin{equation}\label{eq:matrix_m}
	m_i = (\bar{\alpha}_i)^{-1/2} = \begin{cases}
								\sqrt{2} w_i & \text{if} \quad y_i w_i \ge 0 \\
								0		   & \text{else}
							\end{cases}
\end{equation}
and the scalar $t= \sqrt{2}|b|$. For further details can be found in \cite{DBLP:journals/tnn/ChenTY09}.
Even though kernel machines and their derivatives have shown great promise in practical application, their scope is somehow 
limited by the fact that the computational complexity grows rapidly with the size of the kernel matrix (number of data items). 
Among methods suggested to deal with this issue in the literature, the Nystr\"om method has been popular and  widely used.

\section{Nystr\"om approximated matrix processing}
The Nystr\"om approximation technique has been proposed
in the context of kernel methods in \citep{DBLP:conf/nips/WilliamsS00}.
Here, we give a short review of this technique before it is employed in PCVM and iKFD.
One well known way to approximate a $N \times N$ Gram matrix,
is to use a low-rank approximation.
This can be done by computing the eigendecomposition of the kernel matrix
$
{K} = {U} {\Lambda} {U}^T,
$
where ${U}$ is a matrix, whose columns are orthonormal eigenvectors,
and ${\Lambda}$ is a diagonal matrix consisting of eigenvalues 
${\Lambda}_{11} \geq {\Lambda}_{22} \geq ... \geq 0$,
and keeping only the $m$ eigenspaces which correspond to the $m$ largest eigenvalues of the matrix.
The approximation is 
$
{\tilde{K}} \approx {U}_{N,m} {\Lambda}_{m,m} {U}_{m,N},
$
where the indices refer to the size of the corresponding submatrix restricted to the larges $m$ eigenvalues.
The Nystr\"om method approximates a kernel in a similar way,
without computing the eigendecomposition of the whole matrix, which is an $O(N^3)$ operation.

By the Mercer theorem kernels $k(\mathbf{x},\mathbf{x^\prime})$ can be expanded by orthonormal eigenfunctions $\varphi_i$ and non negative eigenvalues $\lambda_i$ in the form
$$
k(\mathbf{x},\mathbf{x^\prime})=\sum_{i=1}^\infty \lambda_i \varphi_i (\mathbf{x}) \varphi_i (\mathbf{x^\prime}).
$$
The eigenfunctions and eigenvalues of a kernel are defined as the solution of
the integral equation
$$
\int k(\mathbf{x^\prime},\mathbf{x}) \varphi_i (\mathbf{x}) p (\mathbf{x}) d\mathbf{x}
= \lambda_i \varphi_i (\mathbf{x^\prime}),
$$
where $p(\mathbf{x})$ is the probability density of $\mathbf{x}$.
This integral can be approximated based on the Nystr\"om technique by an
i.i.d. sample $\{\mathbf{x}^k\}_{k=1}^m$ from $p(\mathbf{x})$:
$$
\frac{1}{m} \sum_{k=1}^m k(\mathbf{x^\prime},\mathbf{x}^k) \varphi_i (\mathbf{x}^k)
\approx \lambda_i \varphi_i (\mathbf{x^\prime}).
$$
Using this approximation we denote with ${K}^{(m)}$ the corresponding
$m \times m$ Gram sub-matrix  and get the corresponding matrix eigenproblem equation as:
$$
{K}^{(m)} {U}^{(m)} = {U}^{(m)} {\Lambda}^{(m)}
$$
with $ {U}^{(m)} \in \mathbb{R}^{m \times m}$ is column orthonormal and 
$ {\Lambda}^{(m)}$ is a diagonal matrix.

Now we can derive the approximations for the eigenfunctions and eigenvalues of the kernel $k$
\begin{equation}
\lambda_i \approx \frac{\lambda_i^{(m)} \cdot N}{m}, \quad  \varphi_i (\mathbf{x^\prime}) \approx \frac{\sqrt{m/N}}{ \lambda_i^{(m)}} \mathbf{k}_x^{\prime,\top} \mathbf{u}_i^{(m)},
\label{eigen-vec_func}
\end{equation}
where $\mathbf{u}_i^{(m)}$ is the $i$th column of ${U}^{(m)}$.
Thus, we can approximate $\varphi_i$ at an arbitrary point $\mathbf{x^\prime}$ as long as we know the vector
$
\mathbf{k}_x^\prime = (k(\mathbf{x}^1,\mathbf{x^\prime}), ... , k(\mathbf{x}^m,\mathbf{x^\prime})).
$
For a given $N \times N$ Gram matrix ${K}$ one may randomly choose $m$ rows and respective columns.
The corresponding indices are called landmarks, and should be chosen such that the,
data distribution is sufficiently covered. 
Strategies how to chose the landmarks have recently been addressed in \cite{DBLP:journals/tnn/ZhangK10a,Zhang} and \cite{DBLP:journals/corr/abs-1303-1849,DeBrabanter20101484}.
We denote these rows by ${K}_{m,N}$.
%
Using the formulas Eq. \eqref{eigen-vec_func} we obtain
$
\tilde{{K}} = \sum_{i=1}^m 1/\lambda_i^{(m)}\cdot
{K}_{m,N}^T (\mathbf{u}_i^{(m)})^T ( \mathbf{u}_i^{(m)} ) {K}_{m,N},
$
where $\lambda_i^{(m)}$ and $\mathbf{u}_i^{(m)}$ correspond to the $m \times m$ eigenproblem.
Thus we get, ${K}^{-1}_{m,m}$ denoting the Moore-Penrose pseudoinverse,
\begin{equation}
	\tilde{{K}}={K}_{N,m} {K}^{-1}_{m,m} {K}_{m,N}.
	\label{Ny_equation}
\end{equation}
as an approximation of ${K}$.
This approximation is exact, if ${K}_{m,m}$ has the same rank as ${K}$.

\subsection{Pseudo Inverse and Singular Value Decomposition of a Nystr\"om approximated matrix}\label{sec:exact_pinv_ny}
In the Ny-PCVM approach discussed in Section \ref{sec:nypcvm} we need the pseudo inverse of a Nystr\"om approximated matrix while for the
Ny-iKFD a Nystr\"om approximated eigenvalue decomposition (EVD) is needed. 

A Nystr\"om approximated pseudo inverse can be calculated by a modified
singular value decomposition (SVD) with a rank limited by $r^* = \min\{r,m\}$ 
where $r$ is the rank of the pseudo inverse and $m$ the number of landmark points.
The output is given by the rank reduced left  and right singular vectors and the reciprocal of the singular values.
The singular value decomposition based on a Nystr\"om approximated similarity matrix $\tilde{K}=K_{Nm} K_{m,m}^{-1} K_{Nm}^\top$ with $m$ landmarks,
calculates the left singular vectors of $\tilde{K}$ as the eigenvectors of $\tilde{K}\tilde{K}^\top$ and the right singular vectors
of  $\tilde{K}$ as the eigenvectors of $\tilde{K}^\top \tilde{K}$\footnote{For symmetric matrices we have
$\tilde{K}\tilde{K}^\top$ = $\tilde{K}^\top\tilde{K}$}. The non-zero singular values of $\tilde{K}$ are then found
as the square roots of the non-zero eigenvalues of both $\tilde{K}^\top\tilde{K}$ or  $\tilde{K} \tilde{K}^\top$. 
Accordingly one only has to calculate a new Nystr\"om approximation of the matrix $\tilde{K} \tilde{K}^\top$
using e.g. the same landmark points as for the input matrix $\tilde{K}$. 
Subsequently an eigenvalue decomposition (EVD) is calculated on the approximated matrix 
$\zeta = \tilde{K} \tilde{K}^\top$.
%
For a matrix approximated by Eq. \eqref{Ny_equation} it is possible to compute
its exact eigenvalue estimators in linear time.
\subsection{Eigenvalue decomposition of a Nystr\"om approximated matrix}
To compute the eigenvectors and eigenvalues of an \emph{indefinite} matrix we first compute the squared form
of the Nystr\"om approximated kernel matrix.
Let ${K}$ be a psd similarity matrix, for which we can write its decomposition as
\begin{align*}
{\tilde{K}}  = {K}_{N,m} {K}^{-1}_{m,m} {K}_{m,N}
& 
= {K}_{N,m} {U} {\Lambda}^{-1}
  {U}^\top {K}_{N,m}^\top
& 
= {B} {B}^\top,
\end{align*}
where we defined ${B}={K}_{N,m} {U} {\Lambda}^{-1/2}$
with ${U}$ and ${\Lambda}$ being the eigenvectors and eigenvalues
of ${K}_{m,m}$, respectively.

\noindent Further it follows for the \emph{squared} ${\tilde{K}}$:
\begin{align*}
{\tilde{K}}^2 = {B} {B}^\top {B} {B}^\top
= {B} {V} {A} {V}^\top {B}^\top,
\end{align*}
where ${V}$ and ${A}$ are the eigenvectors and eigenvalues
of ${B}^\top {B}$, respectively. The square operation
does not change the eigenvectors of $K$ but only the eigenvalues.
The corresponding eigenequation can be written as $B^\top B v = a v$.
Multiplying with $B$ from left we get:
$
	\underbrace{BB^\top}_{\tilde{K}}\underbrace{(B v)}_{u} = a \underbrace{(B v)}_{u} 
$.
It is clear that ${A}$ must be the matrix with the eigenvalues of ${\tilde{K}}$.
The matrix $B v$ is the matrix of the corresponding eigenvectors, which
are orthogonal but not necessary orthonormal.
The normalization can be computed from the decomposition:
\begin{align*}
{\tilde{K}}  = {B} \underbrace{{V} {V}^\top}_{diag(1)} {B}^\top
 = {B} {V} {A}^{-1/2} {A}
  {A}^{-1/2} {V}^\top {B}^\top
= {C} {A} {C}^\top,
\end{align*}
where we defined ${C} = {B} {V} {A}^{-1/2}$
as the matrix of orthonormal eigenvectors of ${K}$.
The eigenvalues of $\tilde{K}$ can be obtained using $A=C^\top \tilde{K} C$. 
Using this derivation we can obtain exact eigenvalues and eigenvectors of an
indefinite low rank kernel matrix $K$, given rank$(K)=m$ and the landmarks
points are independent\footnote{An implementation of this linear time
eigen-decomposition for low rank indefinite matrices is available at: \url{http://www.techfak.uni-bielefeld.de/~fschleif/eigenvalue_corrections_demos.tgz}.}

The former approximation scheme is focused on preserving the full low rank eigen structure of the
underlying data space. The accuracy of this approximation is typically measured by the Frobenius
norm.
A low value of the Frobenius norm of the approximated versus the original kernel matrix ensures
that the approximated kernel matrix $\tilde{{K}}$ can be used similar as $K$ for any kernel
based data analysis method like kernel-PCA, kernel-k-means, SVM, laplacian eigenmaps, preserving
also small between point distances. In the context of classification this requirement is very strong
and unnecessary.  We suggest to restrict the approximation $\tilde{{K}}$ to a low rank kernel
which  preserves \emph{only} the between class distances focusing on class separation.
To achieve this objective we suggest to use a supervised landmark selection scheme. This
is introduced in the following section and compared with a number of baseline methods.


\section{Supervised landmark selection using minimum enclosing balls}
The Nystr\"om approximation is based on $m$ characteristic landmark points taken from the dataset. 
The number of landmarks should be sufficiently large and the landmarks should be diverse enough to get accurate
approximations of the dominating singular vectors of the similarity matrix. In \cite{DBLP:journals/tnn/ZhangK10a}
multiple strategies for landmark selection have been studied and a clustering based approach was suggested
to find the specific landmarks. Thereby the number of landmarks is a user defined parameter and a classical
kmeans algorithm is applied on the kernel matrix to identify characteristic landmark points in the empirical feature space.
This approach is quite effective (see \cite{DBLP:journals/tnn/ZhangK10a}), with some small improvements using an advanced
clustering scheme as shown in \cite{DBLP:conf/icml/SiHD14}. We will use it as a baseline for an advanced landmark
section approach. Further we will also consider a pure random selection strategy. It should be noted that the
formulation given in  \cite{DBLP:journals/tnn/ZhangK10a} takes the full kernel matrix as an input into the k-means
clustering. This is obviously also very costly and may become inapplicable for larger kernel matrices \footnote{
It may however be possible to circumvent this full complexity approach e.g. by subsampling concepts or 
by more advanced concepts of k-means, but this is not the focus of this paper.} 
It is however not yet clear how the \emph{number} of landmarks can be appropriately chosen. If the number of landmarks is large we can expect the data space to be
sufficiently covered after the clustering but the model complexity can become prohibitive. On the other hand
if the number of landmarks is to small the clustering may lead to inappropriate results by merging disjunct parts
of the data space.  We propose to consider the Nystr\"om approximation in a restricted form with respect
to a \emph{supervised} learning problem only. This relieves us from the need of a perfect reconstruction of the kernel
matrix. It is in fact sufficient to reconstruct the kernel such that it is close to the ideal kernel (see e.g. \cite{DBLP:conf/icml/KwokT03}).
We will however not learn an idealized kernel as proposed in \cite{DBLP:conf/icml/KwokT03}, which by itself is
very costly for large scale matrices, but provide a landmark selection strategy leading into a similar direction.

Typically the approximation quality of a Nystr\"om approximated similarity matrix is evaluated using the Frobenius norm.
For real valued data the Frobenius norm of two squared matrices, is simply the sum of the squared difference between the individual kernel entries.
The Frobenius norm is very sensitive to small perturbations of the kernel matrix, addressing also small local
geometric violations. Instead we propose a margin based similarity measure between matrices taking labels into account.

We define a supervised similarity measure between
two matrices to estimate improved prediction performance if a \emph{linear} classifier is used
\footnote{Note that for non-linear separable data the kernel trick can be used and we can still
use linear decision functions as shown in iKFD and PCVM.}
\begin{definition}[Supervised matrix similarity score (SMSS)]\label{def:smss}
	Assume we have a two class problem with $y \in \{-1,1\}$\footnote{The extension to more than two classes is straight forward}. Given a similarity function 
	$K: X \times X \rightarrow \mathbb{R}$ for a learning problem $P$ with
	underlying points $(x,y) \sim P$. We define a score ${s}(\hat{K}, K) = \frac{f(\hat{K})}{f(K)}$ with\\
$
		f(S) = \sum_{y \in \{-1,1\}} \left | E_{(x_i,x^\prime_j) | (y_i=y)} \{ S(x_i,x^\prime_j) | y = y^\prime_j\} \right.
		         \left . -  E_{(x_i,x^\prime_j) | y_i = y}	 \{S(x_i,x^\prime_j) | y \ne y^\prime_j\}\right |
$
\end{definition}
The function $f(S)$ provides a margin estimate of the scores between pairwise similarities of a class $y$ and pairwise
similarities between the entries of class $y$ with respect to entries of the other class(es). Assuming that $f(\hat{K})$ gives 
a margin estimate of the improved matrix and $f({K})$ for the original input matrix, the score  ${s}(\hat{K}, K)>1$, if the margin
has increased and  ${s}(\hat{K}, K)\le 1$ otherwise. Similar as in \cite{DBLP:journals/ml/BalcanBS08}
we assume that a good linear classifier can be obtained from a given similarity function (or kernel) if the similarities between
classes are much lower than those within the same class. However the score ${s}(\hat{K}, K)$ is only a rough estimate. 
It will likely work well for datasets which can easily be modeled by conceptually related classifiers focusing e.g. 
on exemplar based representations. The median classifier proposed in \cite{Nebel2015} is such a simple classifier.
In its simplest form it identifies for each class a single basis function or prototype, showing maximum margin with respect
to the other prototypes with different class labels. In \cite{Nebel2015} it was shown that such a classifier can be very efficient
also for a variety of classification problems. 

In our study the modified similarity matrix $\hat{K}$ is a Nystr\"om approximated matrix, where the landmarks should
be chosen to keep good prediction accuracy instead of a good data reconstruction, as typically aimed for.
Note that due to the approximation it may in fact happen that the SMSS values is below $1$, 
indicating a decreased discrimination power with respect to the full original matrix. But in the considered setting
the approximation is a mandatory step and we try to achieve a large SMSS value for the approximated similarity matrix.

Apparently the (supervised) representation accuracy of the Nystr\"om approximation of $K$ depends on the number and
type of the selected landmarks. We propose to calculate minimum
enclosing ball solutions (MEB) on the individual classwise kernels, to address both problems:
\begin{enumerate}
	\item  finding a sufficient number of landmarks
	\item find landmarks explaining the data characteristics and preserve a good class separation for $\hat{K}$
\end{enumerate}
As an additional constraint we are looking for an approach where also indefinite proximity matrices can
be processed without costly preprocessing steps.
\subsection{MEB for psd input kernels}
We denote the set of indices or points of a sub kernel matrix referring to class $j$ by $R_j$.
Assuming approximately spherical clusters, we can approximate this problem by the
{\bf minimum enclosing ball} :
\[
\begin{array}{ll}
\mbox{min}_{R^2,w_j}&R^2\\
\mbox{such that} &  \|w_j-\Phi(\xi_i)\|^2\le R^2 \quad \forall \xi_i\in R_j
\end{array}
\]
where $R$ is the radius of the sphere and $w_j$ is a center point which
can be indirectly represented in the kernel space) as a weighted linear combination
of the points in $R_j$. The assumption of a sphere is in fact no substantial restriction
if the provided kernel is sufficiently expressive. This is also the reason why core-vector
data description (CVDD) can be used as a linear time replacement for support vector
data description \cite{DBLP:journals/jmlr/TsangKC05}.

It has been shown e.g.\ in 
\cite{DBLP:journals/comgeo/BadoiuC08} 
that the minimum enclosing ball can be approximated
with quality $\epsilon$ in (worst case) linear time using an algorithm
which requires only a constant subset of the receptive field $R_j$, the core set.
Given fixed quality $\epsilon$, the following algorithm
converges in ${\cal O}(1/\epsilon^2)$ steps:

\begin{algorithmic}
\STATE {\bf MEB:}
\STATE $S:=\{\xi_i,\xi_k\}$ for a pair of largest distance $\|\Phi(\xi_i)-\Phi(\xi_k)\|^2$ in $R_j$ and $\xi_i$ chosen randomly\\
\REPEAT
\STATE solve MEB$(S) \to \tilde w_j, R$\\
\IF{exists $\xi_l\in R_j$ where $\|\Phi(\xi_l)-\tilde w_j\|^2>R^2(1+\epsilon)^2$}
\STATE $S:=S\cup\{\xi_l\}$
\ENDIF
\UNTIL{all $\xi_l$ are covered by the $R(1+\epsilon)$ ball in the feature space}
\RETURN $\tilde w_j$
\end{algorithmic}

In each step, the MEB problem is solved for a small subset of constant size only.
This is possible by referring to the dual problem which has the form
\[\begin{array}{ll}
\min_{\alpha_i\ge 0} & \sum_{ij}\alpha_i\alpha_j k_{ij}-\sum_i\alpha_i k_{ii}^2\\
\mbox{where }&\sum_i\alpha_i=1
\end{array}
\]
with data points occurring only as dot products, i.e.\ kernelization is possible.
The same holds for all distance computations of the approximate MEB problem.
Note that the dual MEB problem provides a solution
in terms of the dual variables $\alpha_i$.
The identified finite number of core points (those with non-vanishing $\alpha_i$) will be used as landmarks for this
class and considered to be sufficient to represent the enclosing sphere of the data.
Each class is represented by at least two core points.
Combining all core sets of the various classes provides us with the full set of landmarks
used to get a Nyst\"om approximation of $K$.

The MEB solution typically consists of a very small number 
of points (independent of $N$), sufficient to describe the hyper-ball enclosing the respective data. If the kernel is psd we can use the
MEB approach directly in the kernel space. 

\subsection{MEB for non-psd input kernels}
If the given kernel is non-psd we either can apply various eigenvalue
correction approaches see \cite{Schleif2015f} or we use $\hat{K}=K \cdot K^\top$, which can also be easily done
for Nystr\"om approximated matrices without calculating a full matrix (see first part of Eq. \eqref{eq:squared_mean_reduced_K}). 
This procedure does not change the eigenvectors
of $K$ but takes the square of the eigenvalues such that $\hat{K}$ becomes psd.
It should be noted that if we use 
$\hat{K}$  as an input of a kernel k-means algorithm this is equivalent as using $K$ as the input of the classical
k-means with Euclidean distance as suggested in \cite{DBLP:journals/tnn/ZhangK10a}. 

The proposed supervised landmark selection using MEB does not only identify a good estimate
for the number of landmarks but also ensures that the landmarks are sufficient to explain the data space.
The solutions of the MEB consist of non-redundant points at the perimeter of the sphere, which
can considered to be unrelated, although not necessarily orthogonal in the similarity space
(with potentially squared negative eigenvalues). Especially only those points are included in the
MEB solution which are needed to explain the sphere such that redundancy within this set is
avoided \cite{DBLP:journals/comgeo/BadoiuC08}. Therefore for each
class the MEB solutions provides a local span of the underlying eigen-space. The combination of
the different subspaces can lead to redundancy but we can expect that the full data space
is sufficiently covered. We will show the effectiveness of this approach in some short experiments.

\subsection{Small scale experiments - landmark selection scheme}
\begin{figure}
	\centering
	\begin{subfigure}[b]{0.45\textwidth}
	\includegraphics[width=1\textwidth]{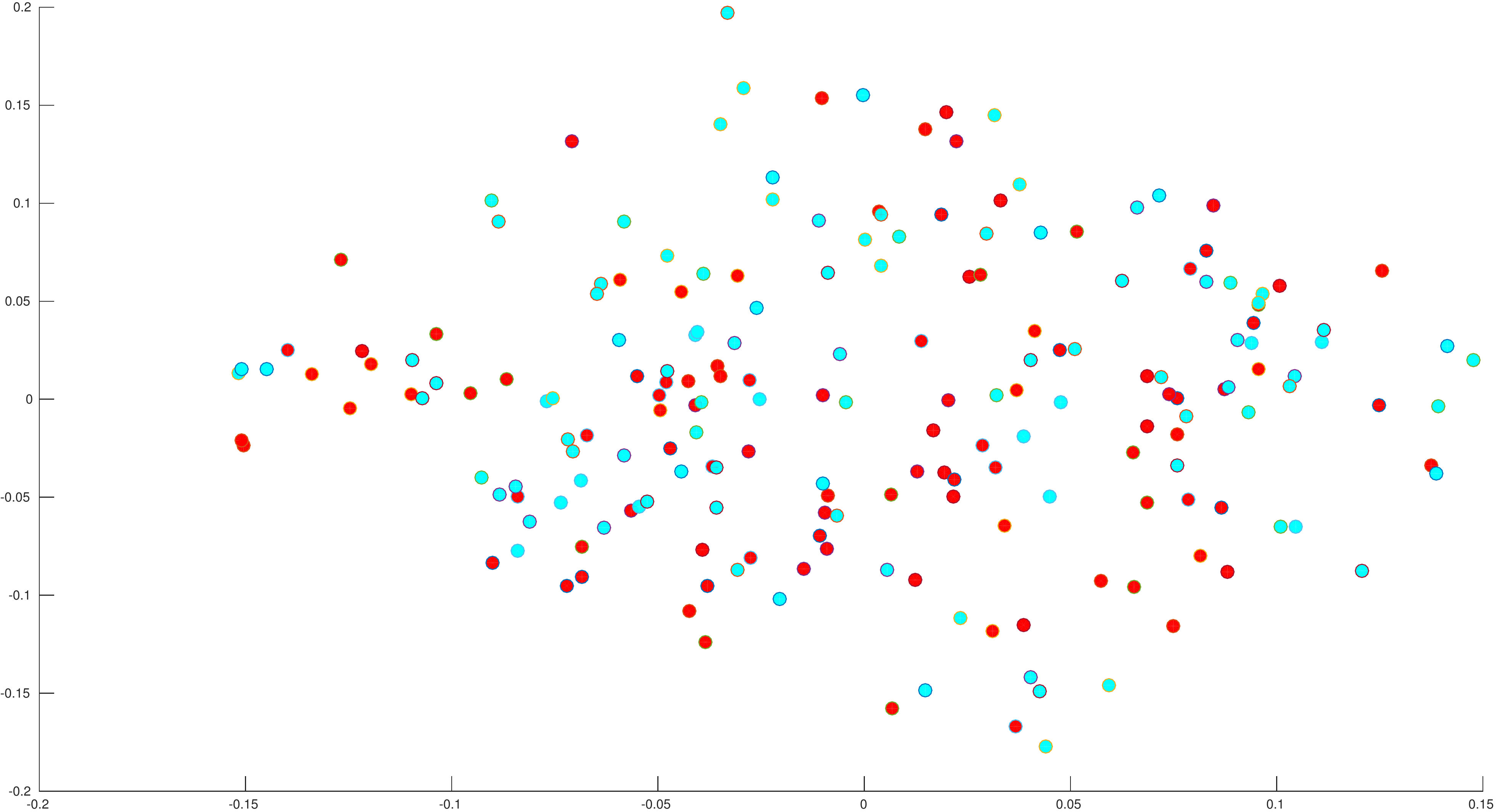}
	\caption{Ball dataset}
	\end{subfigure}
	\begin{subfigure}[b]{0.45\textwidth}
	\includegraphics[width=1\textwidth]{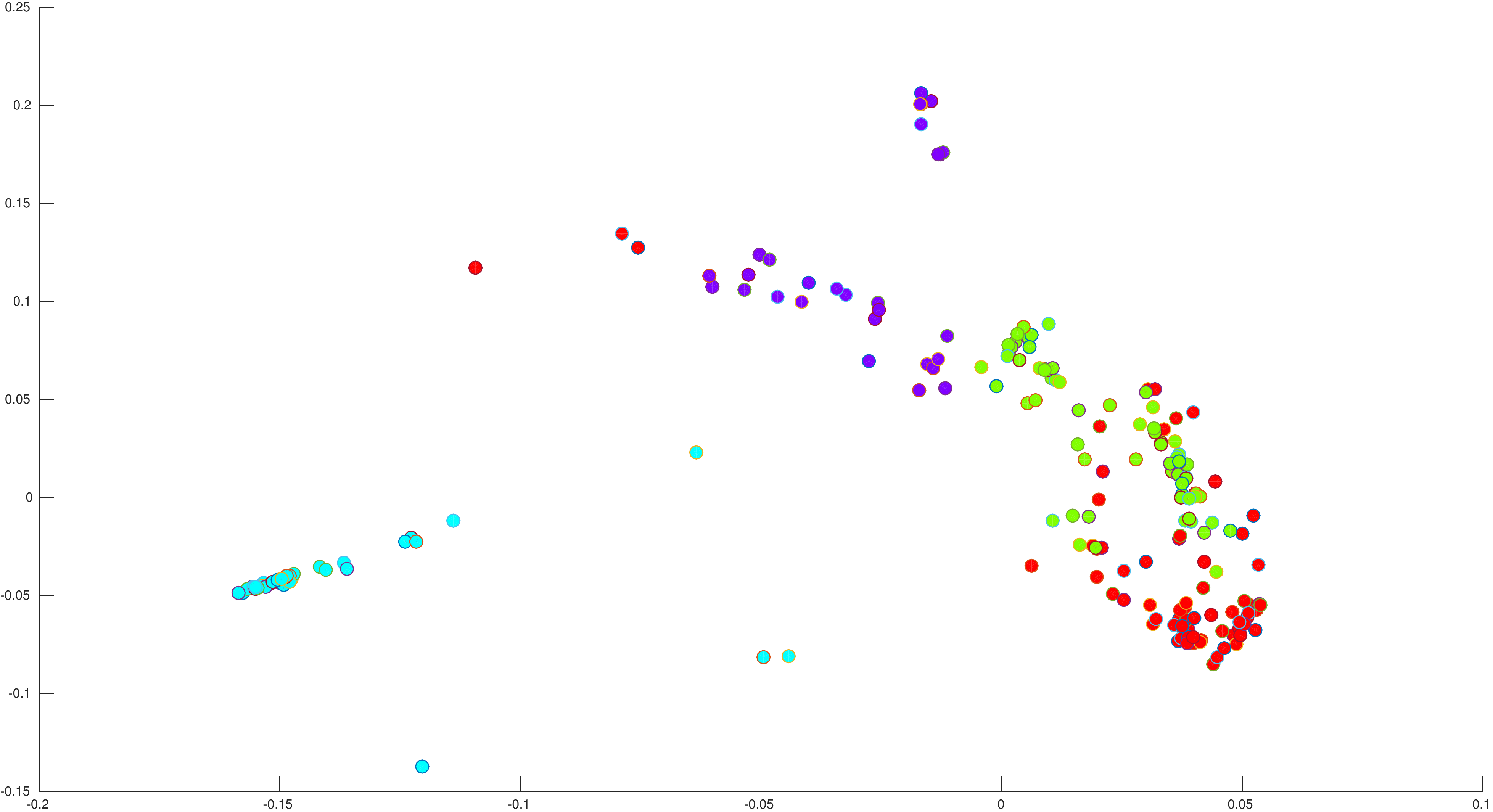}
	\caption{Protein dataset}
	\end{subfigure}
	\begin{subfigure}[b]{0.45\textwidth}
	\includegraphics[width=1\textwidth]{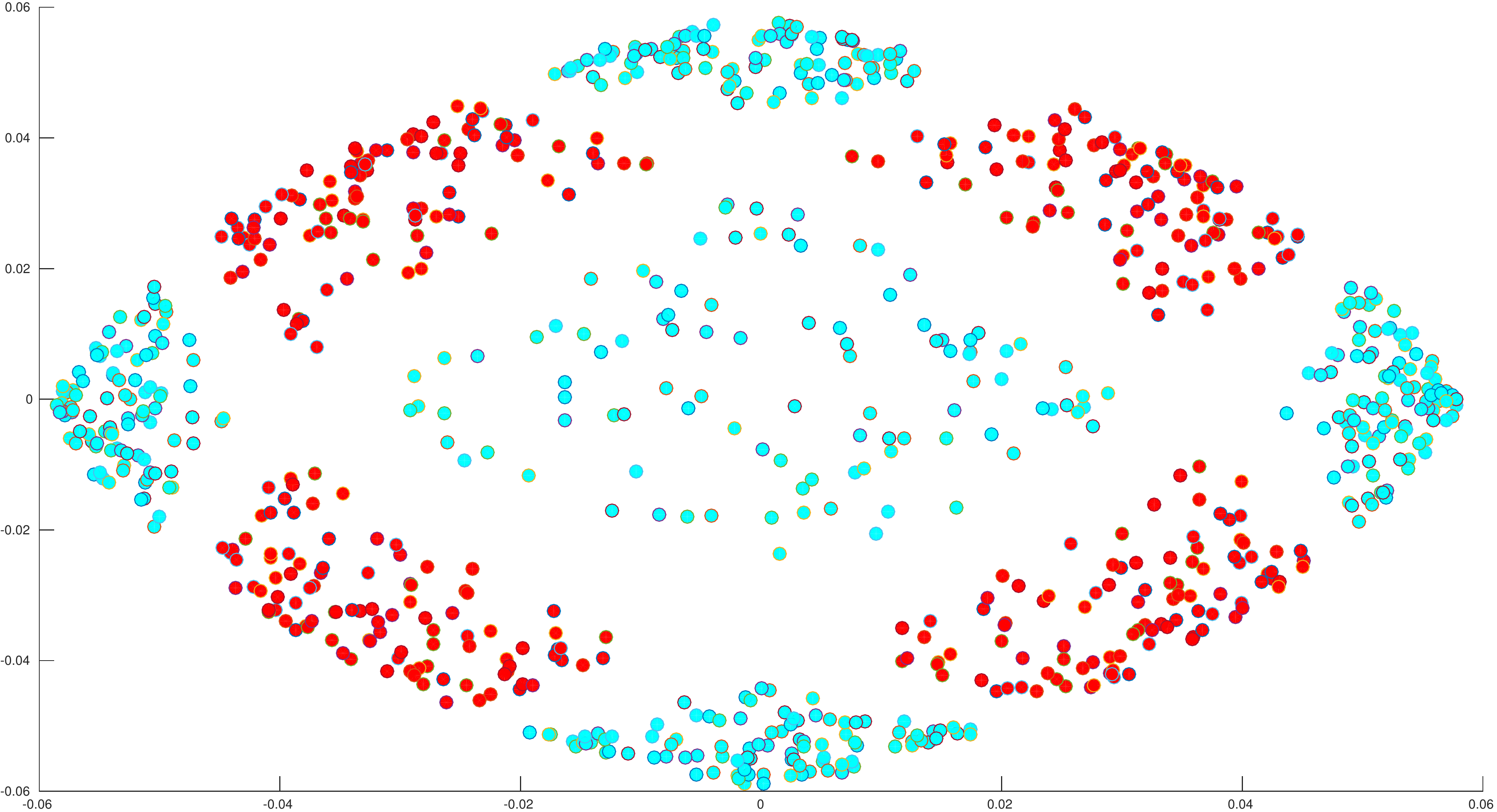}
	\caption{Checker dataset}
	\end{subfigure}
	\begin{subfigure}[b]{0.45\textwidth}
	\includegraphics[width=1\textwidth]{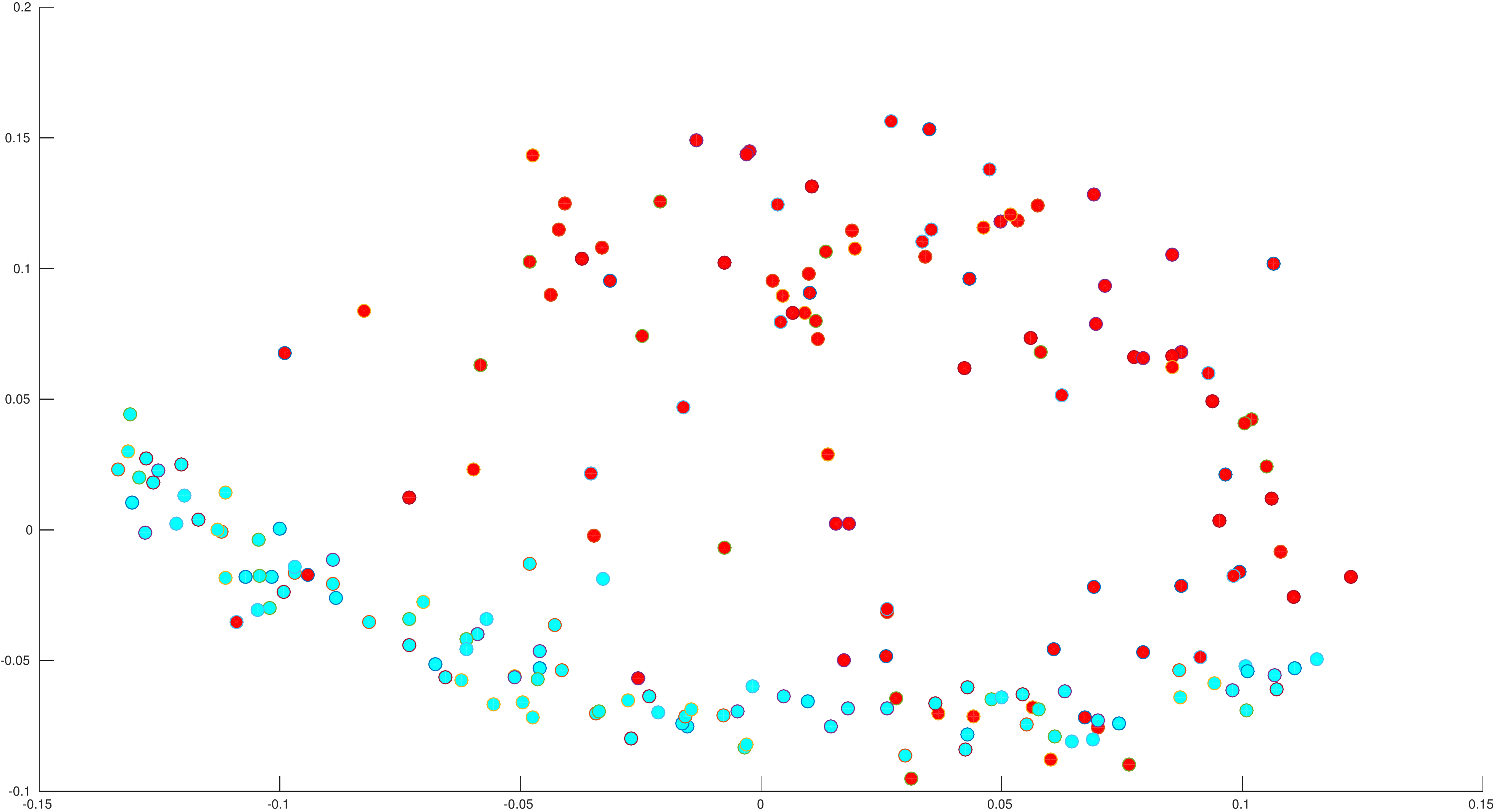}
	\caption{Gauss dataset}
	\end{subfigure}
	\caption{Laplacian eigenmap visualization of the initial test and simulated similarity matrices using $K\cdot K^\top$. Colors/shades indicate the
	different classes. Axis labeling is arbitrary.\label{fig:maps}}
\end{figure}

\begin{figure}
	\centering
	\begin{subfigure}[t]{0.32\textwidth}
	\includegraphics[width=1\textwidth]{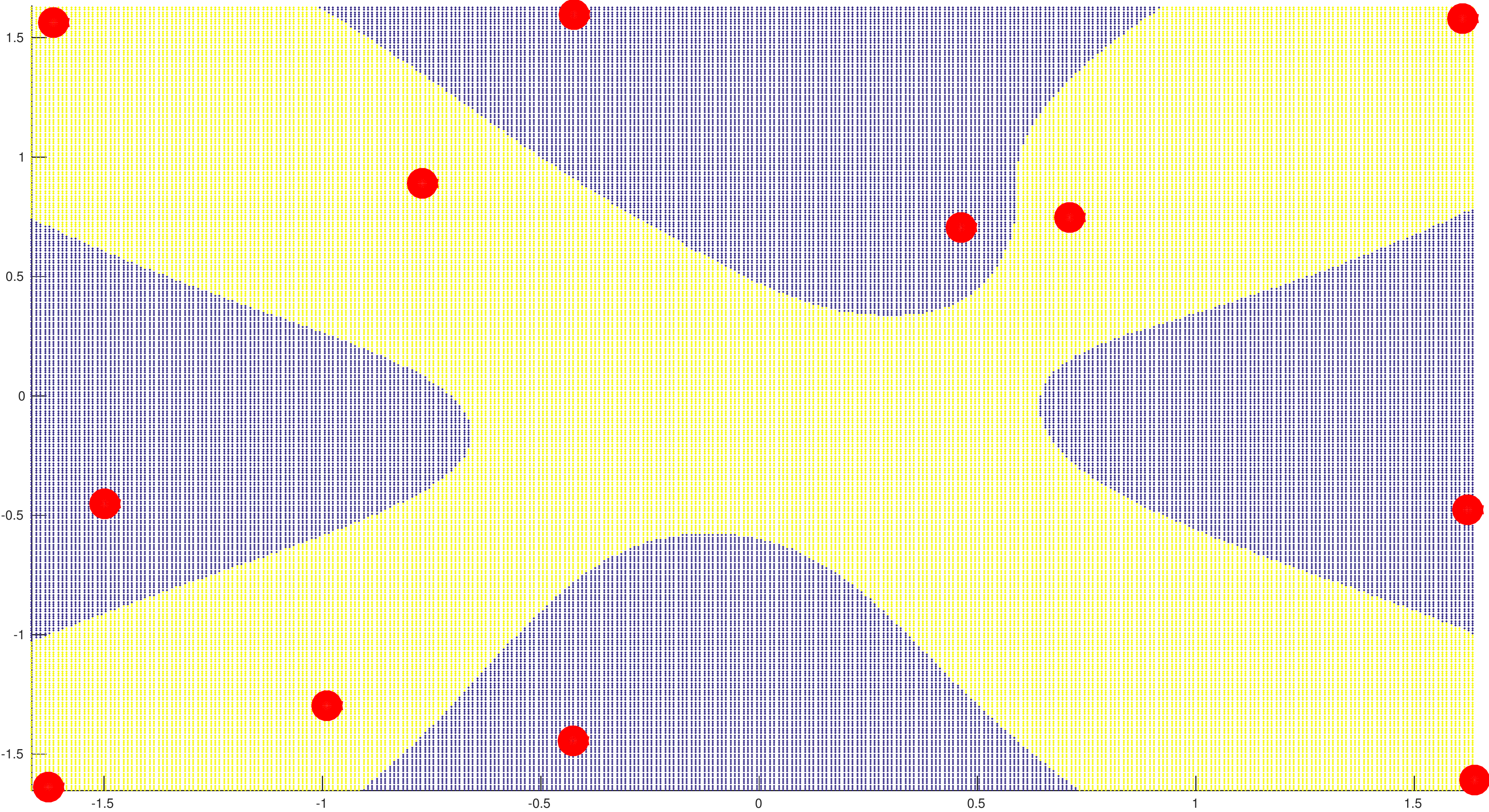}
	\caption{Checker board data with the MEB selection scheme.}
	\end{subfigure}
	\begin{subfigure}[t]{0.32\textwidth}
	\includegraphics[width=1\textwidth]{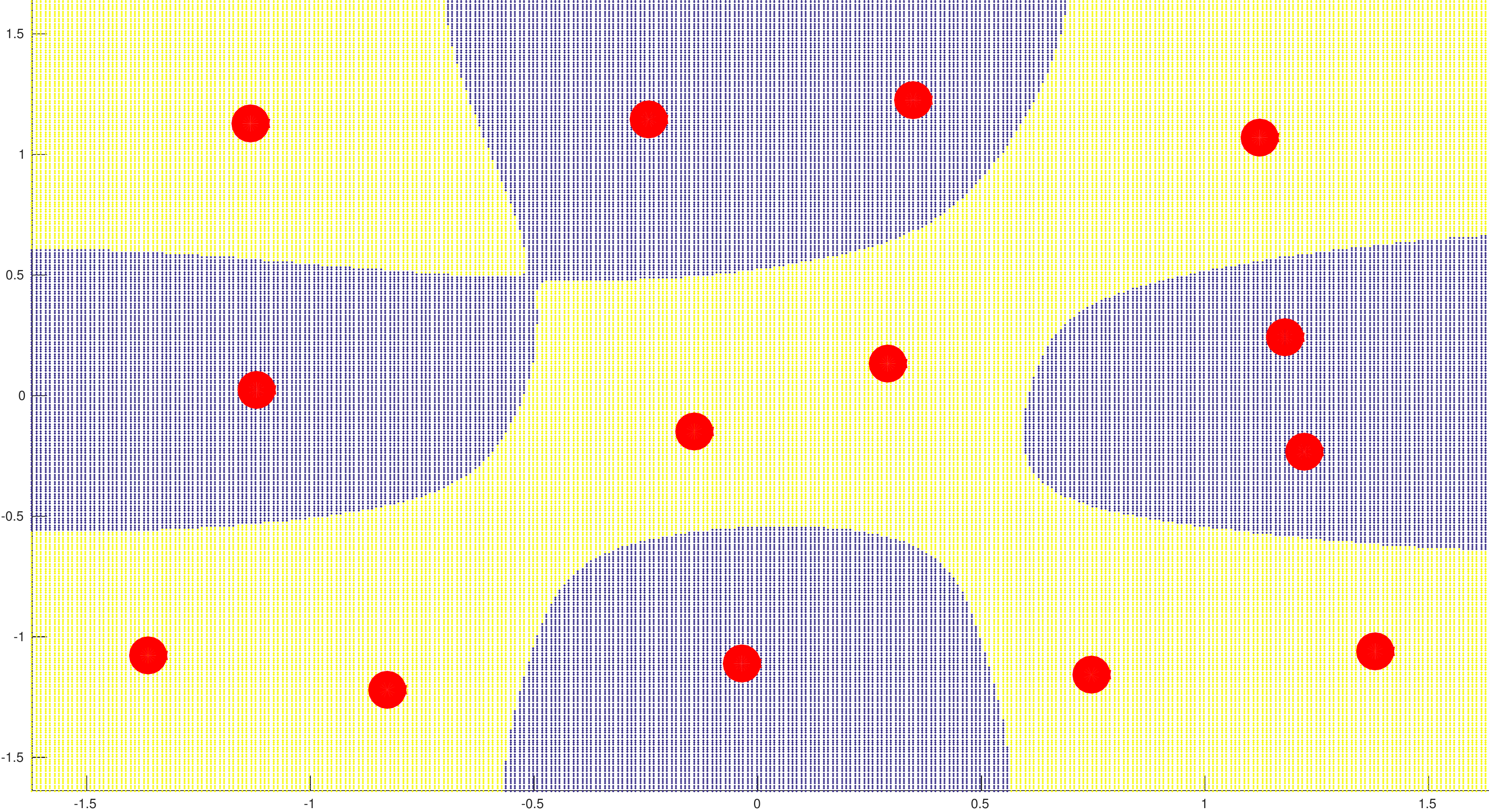}
	\caption{Checker board data with the k-means selection scheme using \#MEB landmarks.}
	\end{subfigure}
	\begin{subfigure}[t]{0.32\textwidth}
	\includegraphics[width=1\textwidth]{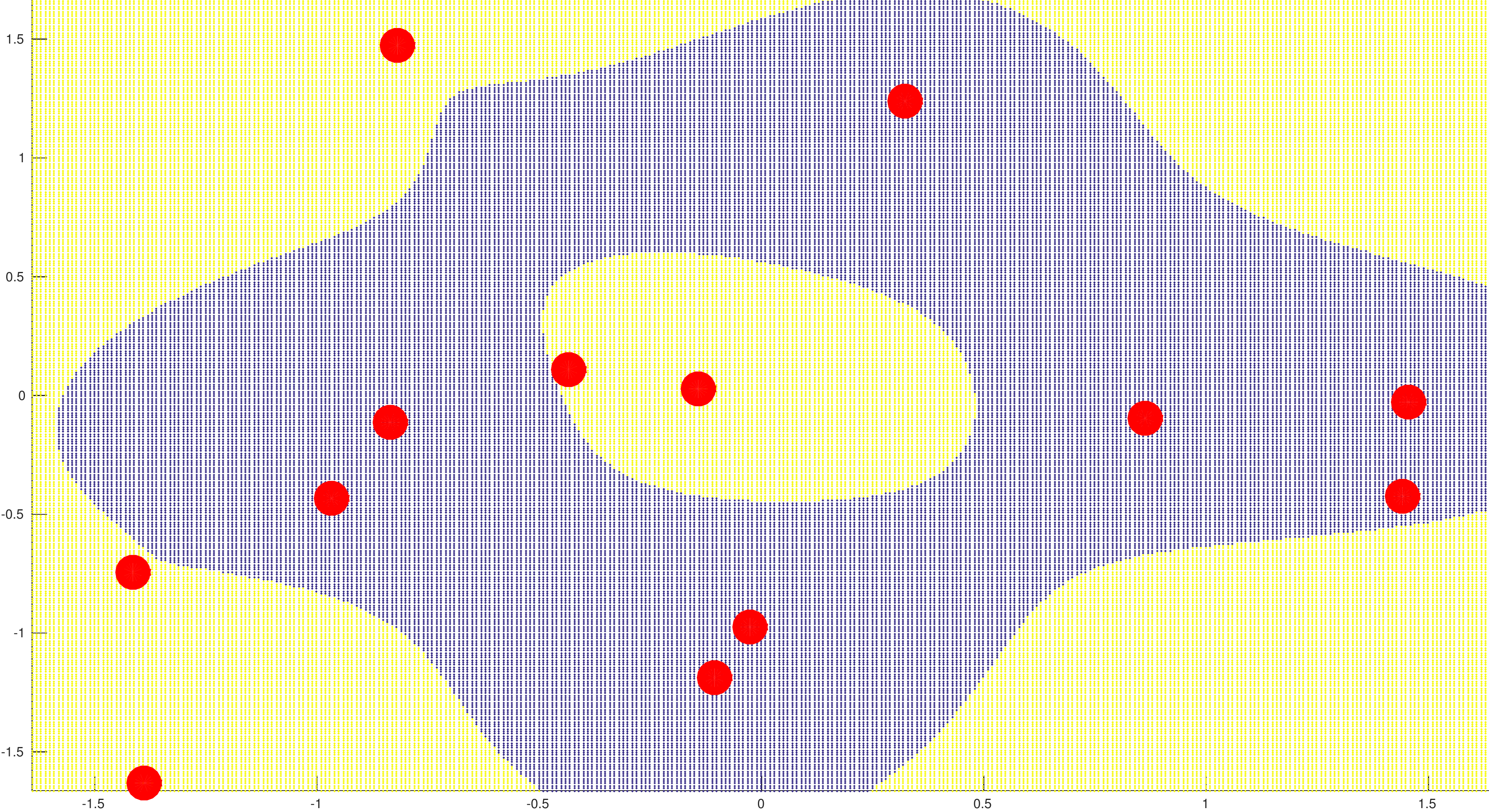}
	\caption{Checker board data with the random selection scheme using \#MEB landmarks.}
	\end{subfigure}
	\caption{Typical plots of the checker board data - taken from the crossvalidation models - with iKFD predictions using different landmark
	selection schemes and an elm kernel. The worst result $\approx 72\%$ is obtained by plot c) using the random
	sampling strategy whereby the number of landmarks was chosen from the MEB approach. The selected landmark points are
	indicated as (red) circles. In plot b) one clearly sees that k-means has rearranged the points to cover the whole data space. For the random
	approach we observe that some points are very close to each other (and have the same label) and are therefore not very informative. \label{fig:checker}}
\end{figure}

We use the ball dataset as proposed in \cite{DBLP:conf/sspr/DuinP10}.
It is an artificial dataset based on the surface distances of randomly positioned balls of two classes having a slightly different radius.
The dataset is non-Euclidean with substantial information encoded in the negative part of the eigenspectrum. 
We generated the data with $100$ samples per class leading to an $N \times N$
dissimilarity matrix $\mathbf{D}$, with $N=200$. 

We also use the protein data (213 pts, 4 classes) set
represented by an indefinite similarity matrix,  with a high intrinsic dimension \cite{Schleif2015f}.
Further we analyzed two simulated metric datasets which are not linear separable using the Euclidean norm:  (1) 
the checker board data, generated as a two dimensional dataset with
datapoints organized on a $3 \times 3$ checkerboard, with alternating labels. This dataset has \emph{multi-modal} classes.
(2) a simple gaussian cloud dataset with two gaussian with substantial overlap. The simulated  data have been represented by 
an extreme learning machine (elm) kernel.  Checker is linear separable in the elm-kernel space, whereas Gaussian is
not separable by construction.

It should be noted that the elm kernel, used for the vectorial data, typically increases the number of non-vanishing eigenvalues such that the original two dimensional data are finally indeed higher dimensional and not representable
by only two basis functions. Two dimensional visualizations of the unapproximated $K\cdot K^\top$ similarity matrices obtained by using laplacian eigenmaps \cite{DBLP:journals/neco/BelkinN03}.
are shown in Figure \ref{fig:maps}.For the checker board data we also show two-dimensional plots of the obtained iKFD decision boundaries and different landmark
selection schemes in Figure \ref{fig:checker}.

Now the obtained (indefinite) kernel matrix has been used in the iKFD in six different ways using different landmark selection schemes: 
\begin{itemize}
	\item[a)] we used the original kernel matrix (SIM1),
	\item[b)] the matrix is Nyst\"om approximated using the MEB approach (SIM2), 
	\item[c)] the matrix is Nystr\"om approximated using the approach of  \cite{DBLP:journals/tnn/ZhangK10a} where the number of landmarks is taken from the MEB solution (SIM3),
	\item[d)] using  the approach of  \cite{DBLP:journals/tnn/ZhangK10a} but with C landmarks where $C$ is the number of classes (SIM4)
	\item[e)] using a random sample of C landmarks (SIM5). SIM5 can be considered as a very basic baseline approach.
	\item[f)] using an entropy based selection as proposed in \cite{DeBrabanter20101484} (SIM6) \footnote{We use the implementation as provided by the authors in the LSSVM toolbox \url{http://www.esat.kuleuven.be/sista/lssvmlab/}}
			where the number of landmarks is again taken from the MEB solution
\end{itemize}

One may also simply use a \emph{very} large number of randomly selected landmarks, but this can become prohibitive if $N$ is large such
that the calculation of $N \times m$ similarities can be costly in memory and runtime. Further it can be very unattractive
to have a larger $m$ for the out of sample extension to new points. If for example costly alignment scores are used
one is interested on having a very small $m$ to avoid large costs in the test phase of the model.

The results of a 10-fold crossvalidation are shown in the Table \ref{tab:sim} with runtimes given in Table \ref{tab:sim_runtimes}. Here and in the following experiments the landmark selection was
part of the crossvalidation scheme and the landmarks are selected on the training set only and the test data have been mapped to the approximated
kernel space by the Nytr\"om kernel expansion (see e.g. \cite{DBLP:conf/nips/WilliamsS00} ).

\begin{figure}
	\centering

	\includegraphics[width=1\textwidth]{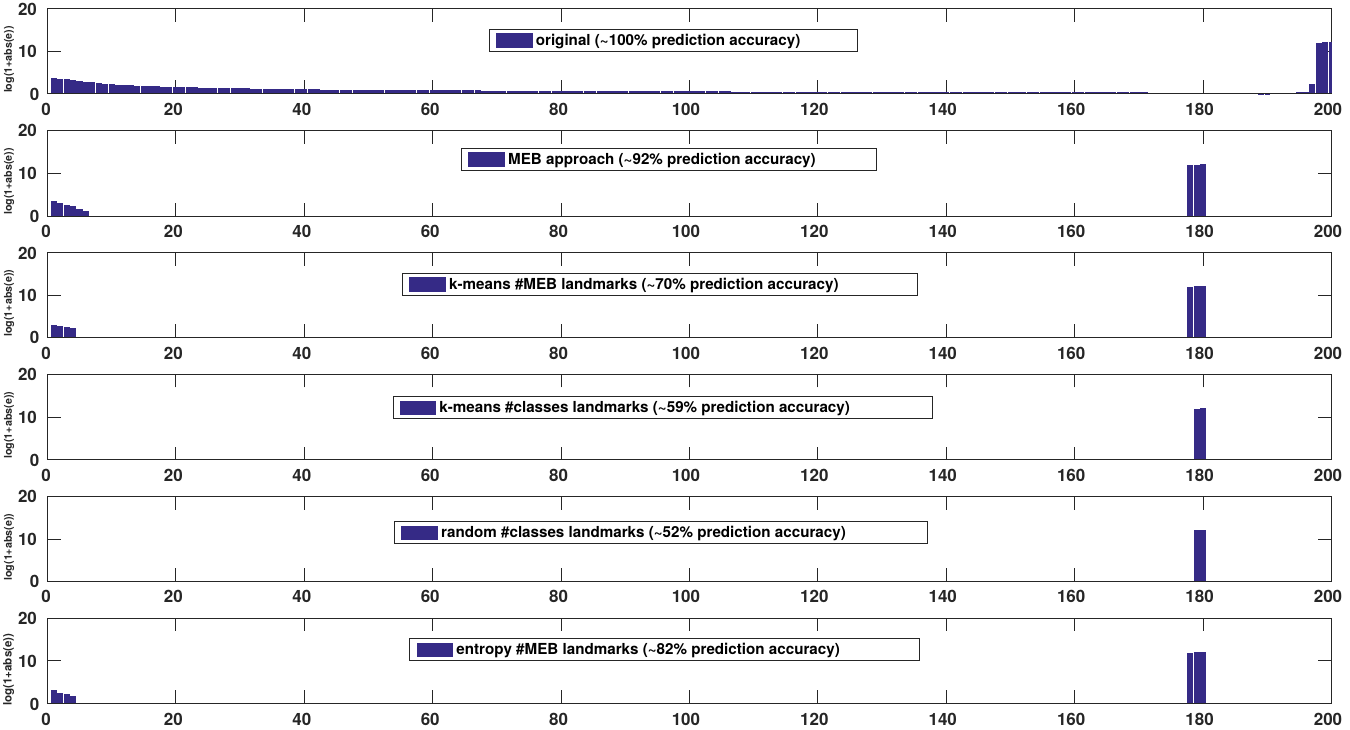}
	\caption{Eigenvalue analysis of the ball dataset using the different approaches. The first plot shows the eigenvalues 
	of the original kernel (SIM1), the other plots show typical results from the 10-fold crossvalidation for the various landmark
	selection approaches (SIM2-SIM6). It can be clearly seen
	that the landmarks identified by the MEB approach sufficiently capture the negative eigenvalues. The random 
	sampling approach works only if a larger number of landmarks is chosen and is still less efficient because it is
	not ensured that the landmarks cover the whole data space. Especially if the data are non i.i.d. random sampling
	is typically insufficient. \label{fig:ball_eigenvalue_analyisis}}
\end{figure}

For the ball data set  the data contain substantial information in the negative fraction of the eigenspectrum, accordingly one may
expect that these eigenvalues should not be removed. This is also reflected in the results. In SIM4 and SIM 5 only the two dominating eigenvectors are kept such that the negative eigenvalues
are removed, degenerating the prediction accuracy. The SIM3 encoding is a bit better, but the landmark optimization via k-means
is not very effective for this dataset. Also the entropy approach in SIM6 was not very efficient. The SIM2 encoding has a substantial drop in the accuracy with respect to the unapproximated kernel
but the intrinsic dimension of the dataset is very high and the $m=8$ landmarks are enough to preserve the dominating positive
and negative eigenvalues.
The unapproximated kernel leads to perfect separation, clearly showing that the negative eigenspectrum contains
discriminative information. The respective eigenvalue plots are provided in Figure \ref{fig:ball_eigenvalue_analyisis}.

The results show that the proposed MEB approach is capable in preserving the geometric information
also for the negative (squared) eigendimensions while being quite simple. We believe that controlling the approximation
accuracy of the kernel by $\epsilon$ in the MEB is much easier than selecting the number of clusters (per class) in k-means clustering.
In fact it will almost always be sufficient to keep $\epsilon \approx 0.01$ to get reliable landmark sets whereas the number
of clusters is very dataset dependent and not easy to choose. However, in contrast to the results shown in Table \ref{tab:sim} the approach 
by  \cite{DBLP:journals/tnn/ZhangK10a} is typically effective for a large variety of datasets also with indefinite kernels, given
the number of landmarks is reasonable large and discriminating information is sufficiently provided in the dominating eigenvectors
of the cluster solutions. For the protein data we observe similar results and the proposed approach, the k-means strategy and the 
entropy approach are effective. SIM4 and SIM5 is again substantially worse because four landmarks are in general not sufficient to represent these
data from a discriminative point of view. 

For the checker board and Gaussian data SIM2 and SIM3 are again close and SIM4 and SIM5 are substantially 
worse using only two landmark points. The entropy approach was efficient only for the Gaussian data, but failed
for Checker which may be attributed to the strong multi-modality of the data. 

\begin{table*}
{\footnotesize
\centering
\caption{\label{tab:sim} Test set results of a 10-fold iKFD run on the simulated / controlled datasets in different kernel approximations
and the obtained SMSS (median) value. 
A $\star$ indicates a non-metric similarity matrix. The number of identified landmarks is shown in brackets for SIM2.}
{\scriptsize

\begin{tabular*}{1\textwidth}{@{\extracolsep{\fill}}l|c|c|c|c}\hline
Method 				& Ball$\star$  				&  		Protein$\star$		&	Checker					&  Gaussian  \\\hline\hline
$SIM1 | {s}(\hat{K}, K)$ 	& $100\pm0 | 1$			& $98.12\pm 3.22 | 1.0$ 			&	$98.89\pm0.35 | 1.0$ 		& $90.00\pm 5.77| 1.0$ \\
$SIM2| {s}(\hat{K}, K)$	& $92.00\pm4.83 | 1.00 (8)$	& $96.71\pm 3.20|0.98 (25) $ 	&	$90.22\pm8.52 | 0.78 (9)$ 	& $90.00\pm 7.45| 0.93 (8)$ \\			
$SIM3| {s}(\hat{K}, K)$	& $70.00\pm12.69 | 0.99$	& $96.71\pm 4.45|0.85 $ 		&	$91.78\pm9.24| 1.0$ 		& $87.00\pm 10.33| 0.98$ \\								
$SIM4| {s}(\hat{K}, K)$	& $59.50\pm5.50 | 0.40$	& $86.85\pm 6.29| 0.71$  		& 	$65.33\pm5.13 | 0.04$		& $65.00\pm 8.17| 0.13$ \\								
$SIM5| {s}(\hat{K}, K)$    & $52.50\pm12.08 | 0.37$	& $78.87\pm14.61 | 1.01$		&      $46.11\pm 4.20| 0.18$ 		& $77.50\pm 10.61| 0.40$ \\
$SIM6| {s}(\hat{K}, K)$    & $74.50\pm12.79|1.00$	& $95.31\pm5.78 | 0.81$  		&	$62.33\pm 11.67|0.10 $ 		& $87.00\pm 7.52| 0.36 $ \\
\end{tabular*}}

}
\end{table*}				

\begin{table*}
{\footnotesize
\centering
\caption{\label{tab:sim_runtimes} Runtimes of a 10-fold iKFD run on the simulated / controlled datasets in different kernel approximations.
A $\star$ indicates a non-metric similarity matrix. }
{\scriptsize
\begin{tabular*}{1\textwidth}{@{\extracolsep{\fill}}l|c|c|c|c}\hline
Method 				& Ball$\star$  				&  		Protein$\star$		&	Checker					&  Gaussian  \\\hline\hline
$SIM1 | {s}(\hat{K}, K)$ 	& $0.5$					& $0.82$			 			&	$13.45$			 		& $0.74$ 					\\
$SIM2| {s}(\hat{K}, K)$	& $1.0$					& $1.56$			 			&	$3.76$			 		& $0.98$ 					\\
$SIM3| {s}(\hat{K}, K)$	& $1.57$					& $2.57$			 			&	$14.77$			 		& $1.51$ 					\\
$SIM4| {s}(\hat{K}, K)$	& $0.84$					& $1.14$			 			&	$13.23$			 		& $0.90$ 					\\
$SIM5| {s}(\hat{K}, K)$    & $0.61$					& $0.98$			 			&	$3.23$			 		& $0.65$ 					\\
$SIM6| {s}(\hat{K}, K)$    & $3.2$					& $8.47$			 			&	$8.12$			 		& $3.94$ 					\\
\end{tabular*}}

}
\end{table*}				

The runtimes shown in Table \ref{tab:sim_runtimes} show already for the small data examples 
that the MEB approach is much faster then k-means or the entropy approach if the number of points gets larger
which was already expected from the theoretical runtime complexity of these algorithms. 

\begin{figure}
	\centering
	\includegraphics[width=0.4\textwidth]{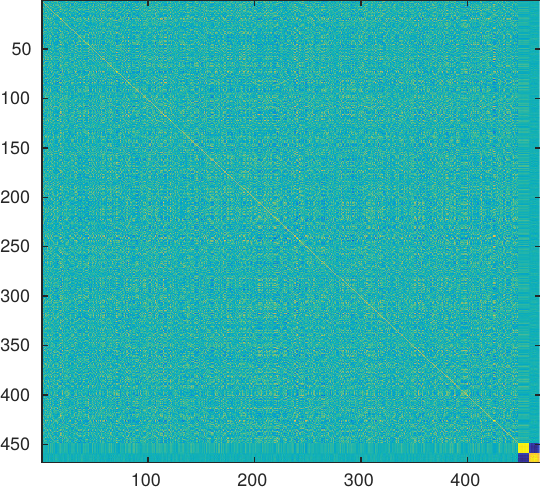}
	\includegraphics[width=0.4\textwidth]{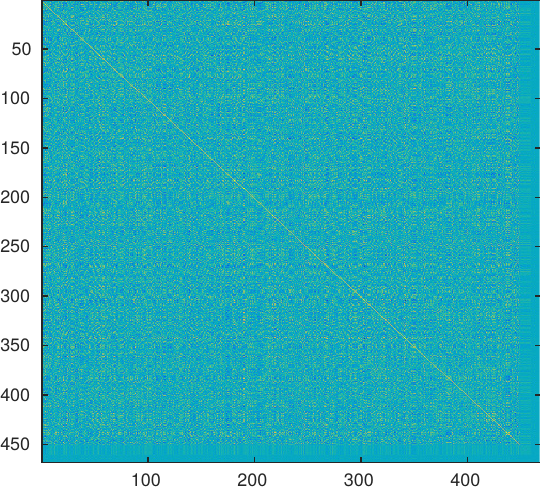}
	\caption{Reconstructed kernel matrix (from the crossvalidation run) of the 10 dimensional Gaussian example. Left using the MEB approach, right 
	using the k-means landmark selection. Note the small  region on the bottom in the left plot indicating the smaller gaussians which are almost missing
	in the right plot.\label{fig:kernel_reconstruction}}
\end{figure}

In another small experiment we analyzed the effect of the k-means based landmark selection  \cite{DBLP:journals/tnn/ZhangK10a}  in more detail.
We consider three Gaussians where one Gaussian has $500$ points spread in two dimensions
and two other Gaussians each with $20$ points spread in another dimensions. All Gaussians are perfectly separated to each other
located in a three dimensional space. To make the task more challenging we further add 7 dimensions with small noise contributions to the large Gaussian.
The final data are given in a 10 dimensional space, whereby the small Gaussians are intrinsically low dimensional and the large Gaussian
is 10 dimensional. with major contributions only in two dimensions. The points from the large Gaussian are labeled 0 and the other 1.
Using the MEB approach we obtain $10$ landmarks and the approximated kernel is sufficient to give a perfect prediction of $100\%$
in a 10-fold crossvalidation with iKFD. Using the k-means or entropy based approach (with the same number of landmarks) the prediction accuracy drops down to $\approx 84\%$
and for random sampling we get a prediction accuracy in the same range of $83\%$ - again with $10$ landmarks .
This can be explained by the behavior of k-means to assign the prototypes or landmarks to dense regions. It is hence more likely
that after the k-means clustering (almost) all prototypes are used to represent the large Gaussian and no prototypes are left for the
other classes. Due to the fact that the other classes are located in different dimensions with respect to the large Gaussian these
dimensions are not any longer well represented and hence the respective classes are often missing in the approximated kernel (see
Figure \ref{fig:kernel_reconstruction}). This density related behavior is also known as magnification \cite{DBLP:journals/neco/VillmannC06}
in the context of different vector quantization approaches. Hence using the unsupervised k-means landmark
selection it can easily happen, that the majority of the data space is well presented but small classes are ignored - which is obviously a
problem for a supervised data analysis.

From these initial experiments we see that the proposed landmark selection scheme is sufficient to approximate the original kernel
function for a \emph{supervised} analysis as indicated by the prediction accuracy of the iKFD model and the SMSS value. We also see that the Nystr\"om approximation can introduce substantial error
if the data are not low rank (for checker) due to a more complicated kernel mapping aka similarity function. We would like
to highlight again that without an advocated guess of the number of landmarks neither the k-means strategy nor the entropy approach
are very efficient. 

\begin{center}
\begin{algorithm*}[tb]
	\begin{enumerate}
		\item let $k(x,y)$ be a symmetric (indefinite) similarity function (e.g. a sequence alignment)
		\item for all labels $c$ let $D_c=\{(x_i,y_i): y_i = c\}$
		\item calculate the (indefinite) kernel matrix $K_c$ using $D_c$ and $k(x,y)$
		\item if the kernel matrix is indefinite, apply a square operation on the small matrix $K_c$ by using $K_c \cdot K_c^\top$
		\item calculate $\forall x \Delta(x) = k(x,x)$ (respectively for all $K_c$)
		\item apply the MEB algorithm for each of the kernel matrices $K_c$ with $\epsilon = 0.01$ and the respective subset of $\Delta$
		\item combine all landmark indices obtained from the former step and calculate the Nystr\"om approximation using Eq. \eqref{Ny_equation}
		\item apply Ny-PCVM or Ny-iKFD using the approximated kernel matrix
	\end{enumerate}
\caption{Proposed handling of indefinite kernels by the MEB approach}
\label{algonyapprox}
\end{algorithm*}
\end{center} 
In the experiment in section \ref{sec:experiments} we will restrict our analysis to the proposed landmark selection using the MEB 
approach, the k-means strategy and the entropy based technique. 

\section{Large scale indefinite learning with PCVM and iKFD}\label{sec:nypcvm} 
We now integrate the aforementioned Nystr\"om approximation approaches and the supervised landmark selection into PCVM 
and iKFD. The modifications ensure that all matrices are processed with linear memory complexity and that the
underlying algorithms have a linear runtime complexity. For both algorithms the initial input is the Nystr\"om approximated kernel
matrix with landmarks selected by using one of the formerly provided landmark selection schemes.

\subsection{PCVM for large scale proximity data}
The PCVM parameters are optimized using the EM algorithm to prune the weight vector $\mathbf{w}$
during learning and hence the considered basis functions representing the model.
We will now show multiple modifications of PCVM to integrate the Nystr\"om approximation 
and to ensure that the memory and runtime complexity remains linear at all time. 
We refer to our method as Ny-PCVM. Initially
the Ny-PCVM algorithm makes use of the matrices $K_1 = {K}_{N,m}$ and  $K_2 = {K}^{-1}_{m,m} \cdot K_1^\top$
obtained from the original kernel matrix using the Nystr\"om landmark technique described above.
Given a matrix 
$X$, we denote by $\hat X$ the matrix formed from $X$  containing elements at indices that have not yet been pruned out of the weight vector $\mathbf w$.
As an example, 
the matrices $\hat{K_1} = K_1^{{\mathbf w}\ne 0,\cdot}$, $\hat{K_2} = K_2^{\cdot,{\mathbf w}\ne 0}$ hold only those columns/rows of $K_1$ or $K_2$ not yet pruned out from the
weight vector. We will use the same notation also for other variables. We denote the set of indices of $m$ randomly selected landmarks by $[m]$. 
Finally, in contrast to the original PCVM
formulation \cite{DBLP:journals/tnn/ChenTY09}, in our notation we explicitly use the data labels - 
for example, instead of vector $\Phi_\theta({\mathbf x})$ we write  $\Xi_\theta({\mathbf x}) \circ {\mathbf y}$,
where $\Xi_\theta({\mathbf x})$ is the kernel vector of ${\mathbf x}$ without any label information, ${\mathbf y}$ is the label vector and $\circ$ is the element-wise multiplication.

We now adapt multiple equations of the original PCVM to integrate the Nystr\"om approximated matrix. 
Beginning with the elements of vector (for a \emph{single} training vector $i$)
 $\mathbf{z}_{\theta}$: 
\begin{equation}
	{z}_{i,\theta}= \Xi_\theta({\mathbf x}_i)  ({{\mathbf y}}\circ {{{\mathbf w}}}) + b,
	\label{z_equation}
\end{equation}
we rewrite Eq.\eqref{z_equation} in matrix notation for all training points:
\begin{equation}
	\mathbf{\hat{z}} = (((\hat{\mathbf{y}} \circ \mathbf{\hat{w}})^\top \hat{K_1}) \cdot K_2)^\top + b
	\label{z_equation_ny}
\end{equation}
and further obtain column vectors $\mathbf{\bar{{H}}_\theta}$ and the reduced form $\mathbf{\bar{\hat{H}}_\theta}$,
by using only the non-vanishing basis functions and the Nystr\"om approximated matrices in Eq. \eqref{eq:decision_function}.
In the maximization step of the original PCVM the $\mathbf{w}$ are updated as (see
Eq. \eqref{eq:weights_update}):
\begin{eqnarray}
	\mathbf{w}^\text{new} &=& M {\underbrace{(M\Phi_\theta({\mathbf x})^\top \Phi_\theta({\mathbf x}) M + I_N)}_{\Upsilon}}^{-1} 
M (\Phi_\theta({\mathbf x})^\top \mathbf{\bar{{H}}_\theta} - b \Phi_\theta({\mathbf x})^\top \mathbf{I})
\end{eqnarray}
To account for the now excluded labels we reformulate Equation \eqref{eq:weights_update} as:
{\small\begin{eqnarray*}
	\mathbf{w}^\text{new} &=& M {\underbrace{(M ({\Xi_\theta({\mathbf x})}^\top {\Xi_\theta({\mathbf x})} {\hat{\mathbf{y}}}^\top{\hat{\mathbf{y}}}) M + I_N)}_{\Upsilon}}^{-1}
M ({\hat{\mathbf{y}}}^\top({\Xi_\theta({\mathbf x})}^\top \mathbf{\bar{H}_\theta}) - b {\hat{\mathbf{y}}}^\top({\Xi_\theta({\mathbf x})}^\top \mathbf{I}))
\end{eqnarray*}}

The update equations of the weight vector include the calculation of a matrix inverse of $\Upsilon$ which 
was originally calculated using the Cholesky decomposition. 
To keep our objective of small matrices we will instead calculate the pseudo-inverse of this matrix 
using a Nystr\"om approximation of $\Upsilon$. It should be noted at this point that the matrix $\Upsilon$
is psd by construction.  We approximate $\Upsilon$ by selecting 
another set of $m^*$ landmarks from the indices of the not yet pruned weights and calculate the matrix
$\tilde{\Upsilon}=C_{Nm^*} W_{m^*,m^*}^{-1} C_{Nm^*}^\top$ in analogy to Eq \eqref{Ny_equation} with submatrices:
\footnote{The number of landmarks $m^*$ is fixed to be $1\%$ of $|w|$ but not more then $500$ landmarks.
If the length of $\mathbf{w}$ drops below $100$ points we use the original PCVM formulations.}
\begin{eqnarray*}
  	C_{Nm^*}        &=& E_{N[m]} + (( \hat{K_1} \cdot (K_2 \cdot (K_1 \cdot \hat{K_2}_{\cdot,[m^*]})) (\hat{\mathbf{y}}^\top\hat{\mathbf{y}}_{[m^*]}) )\\
                                     & &					 \circ \sqrt{2} \mathbf{\hat{\mathbf{w}}}) \circ \sqrt{2} \mathbf{\hat{\mathbf{w}}}_{[m^*]}^\top\\
	W_{m^*,m^*} &=& C_{m^*,\cdot}^{-1}        
\end{eqnarray*}
Where $\circ$ indicates (in analogy to its previous meaning)  that each row of the left matrix is elementwise multiplied by the right vector and
$E_{N[m]}$ is the matrix consisting of the $m$ landmark columns of the $N \times N$ identity matrix. The terms $\sqrt{2} \mathbf{\hat{w}}$ 
and $ \sqrt{2} \mathbf{\hat{w}}_{[m^*]}^\top$ are the
entries of the diagonal matrix $M$ as defined in Eq. \eqref{eq:matrix_m} but now given in 
vector form.

These two matrices serve as the input of a Nystr\"om approximation based pseudo-inverse (as discussed in sub section \ref{sec:exact_pinv_ny}) and
we obtain matrices ${V} \in \mathbb{R}^{N \times r}, {U} \in \mathbb{R}^{r \times N}$ and ${S} \in \mathbb{R}^{r \times  r}$, where $r \le m^*$
is the rank of the pseudo inverse. 
Further we define two vectors $\mathbf{v}_1 = \mathbf{\bar{\hat{H}}_\theta}^\top \cdot K_1$ and $\mathbf{v}_2 = {\mathbf{I}}^\top \cdot K_1$. 
We obtain the approximated weight update
$
	\mathbf{w}^\text{new} = {V} \cdot ({S} \cdot {U}^\top \cdot ( \sqrt{2} \mathbf{\hat{w}} (\hat{\mathbf{y}} (\mathbf{v}_1 \cdot \hat{K_2})^\top
- b \cdot \mathbf{\hat{y}} (\mathbf{v}_2 \cdot \hat{K_2})^\top)))\sqrt{2} \mathbf{\hat{w}}  
$.
The update of the bias is originally done as
\begin{eqnarray}
	\mathbf{b} = t (1+t N t)^{-1} t ( \mathbf{I}^\top   \mathbf{\bar{{H}}_\theta}- \mathbf{I}^\top \Phi_\theta (\hat{\mathbf{y}}   \hat{\mathbf{w}}))
\end{eqnarray}
which is replaced to:
$
	\mathbf{b} = t (1+t N t)^{-1} t (\mathbf{I}^\top   \mathbf{\bar{\hat{H}}_\theta}- \mathbf{I}^\top  ( (((\hat{\mathbf{y}}  \hat{\mathbf{w}})^\top \hat{K_1}) \cdot K_2)^\top   )  )
$
Subsequently the entries in $\mathbf{\hat{w}}$ which are close to zero are pruned out and the matrices $\hat{K_1}$ and $\hat{K_2}$ are modified accordingly.

\subsection{Nystr\"om based Indefinite Kernel Fisher Discriminant}
Given a Nystr\"om approximated kernel matrix a few adaptations have to be made to obtain a valid iKFD formulation
solely based on the Nystr\"om approximated kernel, without any full matrix operations. 

First we need to calculate the classwise means $\mu_+$ and $\mu_-$ based on the row/column sums of the approximated input kernel matrix.
This can be done by rather simple matrix operations on the two low rank matrices of the Nystr\"om approximation of $K$.
For better notation let us define the matrices $K_{Nm}$ as $\Psi$ and $K_{mm}$ as $\Gamma$ then for each row $k$ of the matrix $K$
we get the row/column sum as: 
\begin{eqnarray}\label{eq:row_sum}
	\sum_i [\tilde{K}]_{k,i} = \sum_{l=1}^m \left (\sum_{j=1}^N \Psi_{j,\cdot} \Gamma^{-1} \right ) \Psi_{l,k}^\top
\end{eqnarray}
This can obviously also be done in a single matrix operation for all rows in a batch, with linear complexity only.
Based on these mean estimates we can calculate Eq. \eqref{eq:between_scatter}. In a next step we need to calculate a squared approximated
kernel matrix for the positive and the negative class with removed means $\mu_+$ or $\mu_-$ respectively. For the positive class with $n_+$ entries, we can 
define a new Nystr\"om approximated (squared) matrix with subtracted mean as :
\begin{equation}\label{eq:squared_mean_reduced_K}
	\hat{K}_{N,m}^+ = K_{N,m} \cdot K_{m,m}^{-1} \cdot  ( K_{I_+,m}^\top \cdot K_{I_+,m}) \cdot K_{m,m}^{-1} \cdot K_{m,m}^\top - \mu_+ \cdot \mu_+^\top \cdot n_+
\end{equation}
An equivalent term can be derived for the negative class providing $\hat{K}_{N,m}^-$. It should be noted that no obtained matrix in Eq \eqref{eq:squared_mean_reduced_K}
has more than $N \times m$ entries.
Finally $\hat{K}_{N,m}^+$ and $\hat{K}_{N,m}^-$ are combined to approximate the within class matrix as shown in Eq. \eqref{eq:within_scatter}.
From the derivation in  \cite{Haasdonk2008} we know, that only the eigenvector of the Nystr\"om approximated kernel matrix based on 
$\hat{K}_{N,m} = \hat{K}_{N,m}^+ + \hat{K}_{N,m}^-$ are needed. Using a Nystr\"om based eigen-decomposition (explained before) on $\hat{K}_{N,m}$
we obtain: 
\[
	\alpha = C \cdot A^{-1} \cdot ( C' \cdot (\mu_+-\mu_-))
\]
where $C$ contains the eigenvectors and $A$ the eigenvalues of $\hat{K}_{N,m}$. 
Instead of $A^{-1}$ one can use the pseudo-inverse. The bias term $b$ is obtained as $b = -\alpha^\top (\mu_+ + \mu_-)/2 $.

\section{Complexity analysis}
The original iKFD update rules have costs of $\mathcal{O}(N^3)$ and memory storage $\mathcal{O}(N^2)$, 
where $N$ is the number of points. The Ny-iKFD may involve the extra Nystr\"om approximation of the kernel matrix
to obtain ${K}_{N,m}$ and ${K}_{m,m}^{-1}$, if not already given. If we have $m$ landmarks,
 $m \ll N$, this gives costs of  $\mathcal{O}(mN)$ for the first matrix and $\mathcal{O}(m^3)$ 
for the second, due to the matrix inversion. Further both matrices are multiplied within the optimization so we
get   $\mathcal{O}(m^2 N)$. Similarly, the matrix inversion of the original iKFD with $\mathcal{O}(N^3)$
is reduced to  $\mathcal{O}({m^{2}}N) + \mathcal{O}(m^3)$ due to the Nystr\"om approximation 
of the pseudo-inverse.  If we assume $m\ll N$ the overall runtime and memory complexity of Ny-iKFD is
linear in $N$. For the Ny-PCVM we obtain a similar analysis as shown in \cite{Schleif2015d} but with extra
costs to calculate the Nystr\"om approximated SVD. Additionally, Ny-PCVM uses an iterative optimization
scheme to optimize and sparsify $w$ with constant costs $C_I$, as the number of iterations.
Accordingly Ny-iKFD and Ny-PCVM have both linear memory and runtime complexity $\mathcal{O}(N)$, but
Ny-PCVM maybe slower than Ny-iKFD due to extra overhead costs.

\section{Experiments}\label{sec:experiments}
We compare iKFD, Ny-iKFD, Ny-PCVM and PCVM on various  larger indefinite proximity data. 
In contrast to many standard kernel approaches, for iKFD and PCVM, the indefinite kernel matrices 
need not to be corrected by costly eigenvalue  correction \cite{DBLP:journals/jmlr/ChenGGRC09,Schleif2013b}
\footnote{In \cite{Schleif2015f} various correction methods have been studied on the same data indicating
that eigenvalue corrections may be helpful.}

\begin{figure}
	\centering
	\includegraphics[width=0.9\columnwidth]{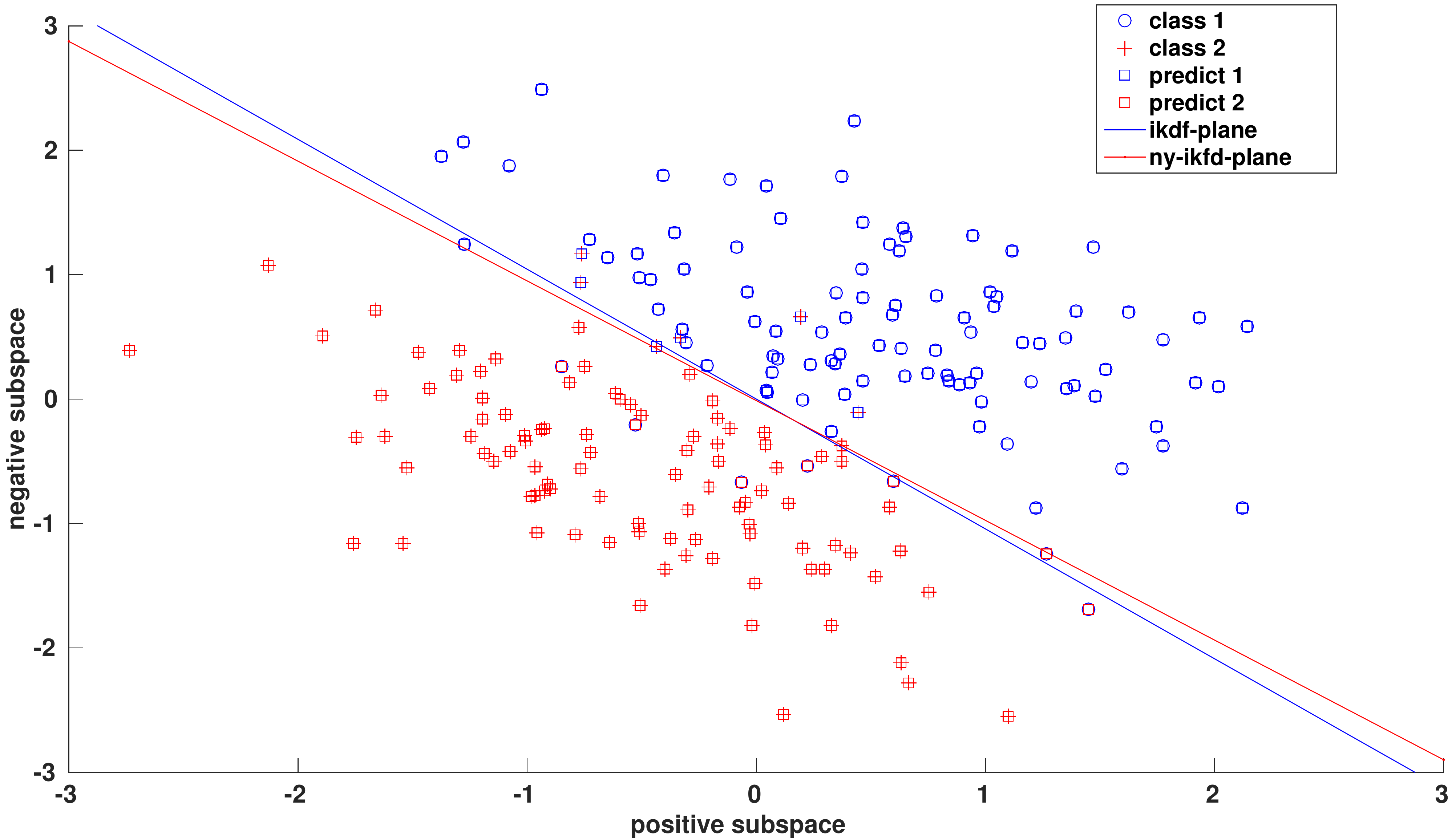}
	\caption{Visualization of the indefinite Fisher kernel for two Gaussians in a two dimensional pseudo-Euclidean space $\mathbb{R}^{(1,1)}$.
	The predicted labels are with respect to the iKFD classification. \label{eq:simulated_example}}
\end{figure}

Further the iKFD and PCVM provides direct access to probabilistic classification decisions.
First we show a small simulated experiment for two Gaussians which exist in an intrinsically two dimensional \emph{pseudo-}Euclidean
space $\mathbb{R}^{(1,1)}$. The plot in Figure \ref{eq:simulated_example} shows a typical result for the obtained
decision planes using the iKFD or Ny-iKFD. The Gaussians are slightly overlapping and both approaches achieve a good 
separation with $93.50\%$ and $88.50\%$ prediction accuracy, respectively. 

Subsequently we consider a few public available datasets for some real life experiments. The data are 
\emph{Zongker} (2000pts, 10 classes) and
\emph{Proteom} (2604pts, 53 classes (restricted to classes with at least 10 entries)) from \cite{PrTools:2012:Online};
\emph{Chromo} (4200pt, 21 classes) from  \cite{neuhaus}
and the SwissProt database \emph{ Swiss} (10988 pts, 30 classes) from \cite{swissprot},
(version 10/2010, reduced to prosite labeled classes with at least $100$ entries ). Further we used the \emph{Sonatas} data (1068pts, 5 classes)
taken from \cite{DBLP:conf/gbrpr/MokbelHH09}.
All data are processed as indefinite kernels and the landmarks are selected using the respective landmark
selection schemes. The mean number of Nystr\"om landmarks as obtained by the MEB approach is given in brackets  after the dataset label.
For all experiments we report mean and standard errors as obtained by a $10$ fold crossvalidation. 
For PCVM we fixed the upper number of optimization cycles to $500$.
The probabilistic outputs can be directly used to allow for a reject region but
can also be used to provide alternative classification decisions e.g. in a ranking framework
\begin{table}[h]
	\small
	\centering
	\begin{minipage}[t]{1\linewidth}
	\begin{tabular*}{1.1\columnwidth}{@{\extracolsep{\fill}} llHllll }\hline
	dataset		 		&      iKFD			  			& iKFD* (MEB)		& Ny-iKFD	(MEB)		&   	 PCVM						& Ny-PCVM (MEB)	  	& $\overline{SMSS}$ \\\hline
	gesture 	(64)			&  $\underline{97.93}\pm0.73$ 	&	$$ 			& $\bf 96.60 \pm 1.84$		&	 $73.20 \pm 18.12$				& $85.53 \pm 1.22*$	&0.9481\\ %
	sonatas 	(25)			&  $90.17\pm2.14$				&	$$ 			& $83.52 \pm 2.08*$		&	 $\underline{91.20} \pm 2.69$	& $\bf 87.08 \pm 3.19*$	&0.7460\\ 
	zongker	(41)		&  $\underline{96.60}\pm1.97$ 	&	$$ 			& $\bf 90.70 \pm 2.30*$		&  	 $93.60 \pm 2.00$				& $84.35 \pm 2.53*$ 	&0.6785\\ 
	proteom	(123)		& $99.58\pm0.38$ 				&	$$ 			& $\bf \underline{99.68} \pm 0.31$&	 $99.58 \pm 0.28$				&  $99.45 \pm 0.53$	&0.9184\\ 
	chromo	(65)			& $\underline{97.24}\pm0.94$ 	&	$$ 			& $\bf 94.79 \pm 1.45$	 	&  	 $93.29 \pm 1.51$				& $92.21 \pm 1.31$ 	&0.9300\\ 
	swiss	(116)       		&     --  						& -				&$\underline{83.05} \pm 1.60$			&  	--							&  $70.38 \pm 19.19$  	&0.7870\\\hline %
	\end{tabular*}  
	\caption{Crossvalidation results using the MEB approach.  iKFD and PCVM use the original
	indefinite kernel without approximations. Ny-iKFD and Ny-PCVM use the Nystr\"om approximation within the implementation as discussed before and the same Nystr\"om approximated
	kernel. (*) indicate significant differences with respect to the same unapproximated method.  The mean $\overline{SMSS}$ values are calculated on the MEB based Nystr\"om approximation.\label{tab:large_npsd_acc_core}}
	\end{minipage}
\end{table}
\begin{table}[h]
	\small
	\centering
	\begin{minipage}[t]{1\linewidth}
	\begin{tabular*}{1.1\columnwidth}{@{\extracolsep{\fill}} llHllll }\hline
	dataset		 		&      iKFD			  			& iKFD* (KM)			& Ny-iKFD (KM)		&   	 PCVM								& Ny-PCVM 	   (KM)	& $\overline{SMSS}$ \\\hline
	gesture  (64)			&  $\underline{97.93}\pm0.73$ 	&	$$ 				& $95.73 \pm 0.86$	&	 $73.20 \pm 18.12$						& $92.60 \pm 1.04*$	& 0.9315\\ %
	sonatas 	(25)			&  $90.17\pm2.14$				&	$83.52\pm4.77$ 	& $77.63 \pm 3.19*$	&	 $\underline{91.20} \pm 2.69$			& $77.81 \pm 3.28*$	& 0.4925\\ 
	zongker	(41)		&  $\underline{96.60}\pm1.97$ 	&	$91.85\pm2.27$ 	& $88.40 \pm 1.33*$	&  	 $93.60 \pm 2.00$						& $88.30 \pm 2.89*$ 	& 0.8340\\ 
	proteom	(123)		& $99.58\pm0.38$ 				&	$93.39\pm0.68$ 	& $94.78 \pm 1.89$	&   	 $99.58 \pm 0.28$						&  $94.18 \pm 1.23$	& 0.5711\\ 
	chromo	(65)			& $\underline{97.24}\pm0.94$ 	&	$94.98\pm1.07$ 	& $ 94.17 \pm 0.86$ 	&  	 $93.29 \pm 1.51$						& $92.10 \pm 0.89$ 	& 0.9406\\ 
	swiss	(116)       		&     --  						& -					&$73.74 \pm 0.71$		&  	--									&  $75.36 \pm 7.55$  	& 0.7864\\\hline %
	\end{tabular*}  
	\caption{Crossvalidation results using the k-means (KM) approach.	iKFD and PCVM use the original indefinite kernel without approximations. 
	Ny-iKFD and Ny-PCVM are Nystr\"om approximated and use the same Nystr\"om approximated kernel obtained by the KM strategy.  
         (*) indicate significant differences with respect to the same unapproximated method. The number of landmarks is chosen w.r.t. the MEB solution.
	The mean $\overline{SMSS}$ values are calculated on the KM based Nystr\"om approximation.\label{tab:large_npsd_acc_kmeans}}
	\end{minipage}
\end{table}
\begin{table}[h]
	\small
	\centering
	\begin{minipage}[t]{1\linewidth}
	\begin{tabular*}{1.1\columnwidth}{@{\extracolsep{\fill}} llHllll }\hline
	dataset		 		&      iKFD			  			& iKFD* (ENT)			& Ny-iKFD	(ENT)		&   	 PCVM					& Ny-PCVM (ENT)	  	& $\overline{SMSS}$ \\\hline
	gesture 	(64)			&  $97.93\pm0.73$				&	$94.00\pm1.78$	& $ 93.47\pm 1.93$*	&	 $73.20\pm18.12$				& $91.07\pm2.97$*	& 0.8864\\ %
	sonatas 	(25)			&  $90.17\pm2.14$				&	$\pm$ 			& $ 80.24\pm 2.46$*	&	 $91.20\pm2.69$				& $82.77\pm2.86$*	& 0.5210\\ 
	zongker	(41)			&  $96.60\pm1.97$				&	$\pm$ 			& $ 90.90\pm 1.15$*	&	 $93.60\pm2.00$				& $90.50\pm2.12$		& 0.5500\\ 
	proteom	(123)		&  $99.58\pm0.38$				&	$\pm$ 			& $ 94.54\pm 1.87$	&	 $99.58\pm0.28$				& $80.93\pm22.96$*	& 0.5057\\ %
	chromo	(65)			&  $97.24\pm0.94$				&	$94.43\pm0.64$*	& $ 94.50\pm 1.30$	&	 $93.29\pm1.51$				& $90.95\pm2.55$		& 0.9560\\ 
	swiss	(116)       		&  - 							&	$\pm$ 			& -					&	-							& -					& \\\hline %
	\end{tabular*}  
	\caption{Crossvalidation results using the entropy (ENT) approach.
	Ny-iKFD and Ny-PCVM are Nystr\"om approximated and use the same Nystr\"om approximated kernel obtained by the entropy strategy.  
         (*) indicate significant differences with respect to the same unapproximated method. The number of landmarks is chosen w.r.t. the MEB solution.
	The mean $\overline{SMSS}$ values are calculated on the ENT based Nystr\"om approximation. For the swiss data the entropy approach took to much time.\label{tab:large_npsd_acc_ent}}
	\end{minipage}
\end{table}

\begin{table}[h]
	\begin{minipage}[t]{1\linewidth}
	\begin{tabular*}{1\columnwidth}{@{\extracolsep{\fill}} lllll }\hline
				 		&   iKFD 					& Ny-iKFD 			&  PCVM				& Ny-PCVM  \\\hline
	gesture				&   $50.72\pm1.54 $ 	& $9.18 \pm 0.19$			&  $116.33 \pm 7.49 $ 	& $31.98\pm0.42$\\
	sonatas 				&  $5.04\pm0.22$		& $1.85\pm 0.06$			&  $60.07 \pm 2.54$	& $7.01\pm0.24$	\\
	zongker				&  $51.61\pm1.43 $	& $5.53 \pm 0.16$			&  $184.07\pm 14.97$	&  $16.91\pm0.24$\\ 
	proteom				& $559.25\pm15.29$ 	& $42.08 \pm 1.92$ 		&  $352.08 \pm 18.05$	&  $111.22\pm1.88$ \\ 
	chromo				&  $763.24\pm31.54$ 	&$27.91 \pm 1.77$			&  $694.43 \pm 15.61$ 	&  $54.36\pm0.77$ \\ 
	swiss	       			&-- 			 		&$178.79 \pm 10.63$		&  -- 					&    $123.29\pm2.72$	\\\hline 
	\end{tabular*}  
	\caption{Typical runtimes - indefinite kernels\label{tab:large_npsd_time}}	
	\end{minipage}
\end{table}
			
\begin{table}[h]
	\begin{minipage}[t]{1\linewidth}
	\begin{tabular*}{1\columnwidth}{@{\extracolsep{\fill}} lllll }\hline
				 		&   iKFD 					& Ny-iKFD (MEB)			&  PCVM				& Ny-PCVM  (MEB)\\\hline
	gesture				&   $100.00\pm0$		& $100.00\pm0$				&  $ 10.60 \pm 0.84$	& $5.25 \pm 0.31$\\
	sonatas 				&  $100.00\pm0$ 		& $100.00\pm 0$				&  $ 11.24\pm 0.56 $	& $3.42\pm0.57$\\
	zongker				& $100.00\pm0$ 		& $100.00\pm0 $				&  $ 14.42\pm 3.65 $	& $8.63 \pm 0.31$\\
	proteom				&  $100.00\pm0$ 		& $100.00\pm0 $				&  $ 5.23\pm 0.36 $ 	& $5.85 \pm 0.14$\\
	chromo				&  $100.00\pm0$ 		& $100.00\pm0$				&  $ 7.49\pm 0.51 $ 	& $2.49 \pm 0.34$\\
	swiss	       			&$-$ 				& $96.95 \pm 0.27$			&  $ -$ 				& $1.18 \pm 0.25$\\\hline
	\end{tabular*}  
	\caption{Model complexity - indefinite kernels (threshold $1e^{-4}$) \label{tab:complexity}}	
	\end{minipage}
\end{table}
	
In Table \ref{tab:large_npsd_acc_kmeans}, \ref{tab:large_npsd_acc_core} and Table \ref{tab:large_npsd_time} we show the results for different
non-metric proximity datasets using Ny-PCVM, PCVM and iKFD or Ny-iKFD. The overall best results for a 
dataset are underlined and the best approximations are highlighted in bold.
Considering Table \ref{tab:large_npsd_acc_kmeans} and Table \ref{tab:large_npsd_acc_core}
we see that iKFD and PCVM are similarly effective with slightly better results for iKFD. The Nystr\"om approximation
of the kernel matrix \emph{only}, often leads to a in general small decrease of the accuracy, but
the additional approximation step, in the algorithm itself, does not substantially decrease the
prediction accuracy further\footnote{Also the runtime and model complexity are similar and therefore not reported in the following.}. 

Considering the overall results in Table  \ref{tab:large_npsd_acc_kmeans} and Table  \ref{tab:large_npsd_acc_core} 
the approximations used in the algorithms Ny-iKFD and Ny-PCVM appear to be effective.
The runtime analysis in Table  \ref{tab:large_npsd_time} clearly shows that the classical iKFD is very complex.
As expected, the integration of the Nystr\"om approximation leads to substantial speed-ups. Larger datasets like the
Swiss data with $\approx 10.000$ entries could not be analyzed by iKFD or PCVM before. We also see that the
landmark selection scheme using MEB is slightly more effective than by using k-means but without the need to tune
the number of clusters (landmarks). The entropy approach is similar efficient than the k-means strategy but
more costly due to the iterative optimization of the landmark set and the respective eigen-decompositions (see \cite{DeBrabanter20101484}).

The PCVM is focusing on a sparse parameter vector $w$ in contrast to the iKFD. For the iKFD
most training points are also used in the model $(\ge 94\%)$ whereas for Ny-PCVM 
often less than $5\%$ are kept in general as shown in Table \ref{tab:complexity}.
In practice it is often costly to calculate the non-metric proximity measures like sequence
alignments and also a large number of kernel expansions should be avoided. Accordingly sparse models are very desirable. 
Considering the runtime again Ny-PCVM and Ny-iKFD
are in general faster than the original algorithms, typically by at least a magnitude. the PCVM and Ny-PCVM are 
also very fast in the test case or out-of sample extension due to the inherent model sparsity.

\section{Conclusions}
We presented an alternative formulation of the iKFD and PCVM
employing the Nystr\"om approximation. We also provided an alternative way to identify the landmark points 
of the Nystr\"om approximation in cases where the objective is a supervised problem. Our results indicate 
that in general the MEB approach is similar efficient compared to the k-means clustering or the entropy strategy 
but with less effort and almost parameter free.
We found that Ny-iKFD is competitive in the prediction accuracy
with the original iKFD and alternative approaches, while taking substantially less memory and runtime but
being less sparse then Ny-PCVM.  The Ny-iKFD and Ny-PCVM provides now an effective 
way to obtain a \emph{probabilistic} classification model for medium to large psd \emph{and} non-psd datasets,
in \emph{batch mode} with \emph{linear} runtime and memory complexity. If sparsity is not an issue one may prefer Ny-iKFD
which is slightly better in the prediction accuracy then Ny-PCVM.
Using the presented approach we believe
that iKFD is now applicable for realistic problems and may get a larger impact then before. In future work it could
be interesting to incorporate sparsity concepts into iKFD and Ny-iKFD similar as shown for classical KFD in \cite{DBLP:journals/jmlr/DietheHHS09}.
\\[1cm]
{{\bf Implementation:}}
The Nystr\"om approximation for iKFD is provided at \url{http://www.techfak.uni-bielefeld.de/~fschleif/source/ny_ikfd.tgz}
and the PCVM/Ny-PCVM code can be found at \url{https://mloss.org/software/view/610/}.
\\[1cm]
{{\bf Acknowledgment:}
A Marie Curie Intra-European Fellowship (IEF): FP7-PEOPLE-2012-IEF (FP7-327791-ProMoS) and
support from the Cluster of Excellence 277 Cognitive Interaction Technology funded by the 
German Excellence Initiative is gratefully acknowledged. PT was supported by the EPSRC grant EP/L000296/1, 
"Personalized Health Care through Learning in the Model Space". We would like to thank R. Duin, Delft University for 
various support with distools and prtools and Huanhuan Chen,University of Science and Technology of China, 
 for providing support with the Probabilistic Classification Vector Machine.}


\end{document}